\documentclass{article}
\PassOptionsToPackage{numbers, compress}{natbib}

\usepackage[main, final]{neurips_2026}

\usepackage[utf8]{inputenc} 
\usepackage[T1]{fontenc}    
\usepackage{hyperref}       
\usepackage{url}            
\usepackage{booktabs}       
\usepackage{amsfonts}       
\usepackage{nicefrac}       
\usepackage{microtype}      
\usepackage{xcolor}         
\usepackage{graphicx}
\usepackage{amsmath}
\usepackage{amssymb}
\usepackage{multirow}
\usepackage{algorithm}
\usepackage{algorithmic}
\usepackage{wrapfig}
\usepackage{makecell}
\usepackage{caption}
\usepackage[normalem]{ulem}

\title{GCD: GCM-consistent Diffusion for Zero-shot Downscaling across Heterogeneous GCMs}

\author{%
  \textbf{Ruian Tie}\textsuperscript{\textnormal{1,2,3,*}} \quad
  \textbf{Wenbo Xiong}\textsuperscript{\textnormal{4,2,3,*}} \quad
  \textbf{Zhengyu Shi}\textsuperscript{\textnormal{1,3,*}} \quad
  \textbf{Xinyu Su}\textsuperscript{\textnormal{1,3}} \\
  \textbf{Chenyu Jiang}\textsuperscript{\textnormal{4,3}} \quad
  \textbf{Libo Wu}\textsuperscript{\textnormal{5,2,3,\textdagger}} \quad
  \textbf{Hao Li}\textsuperscript{\textnormal{1,2,3,\textdagger}} \\
  \textsuperscript{1} Artificial Intelligence Innovation and Incubation Institute, Fudan University, Shanghai, China \\
  \textsuperscript{2} Shanghai Innovation Institute, Shanghai, China \\
  \textsuperscript{3} Shanghai Academy of AI for Science, Shanghai, China \\
  \textsuperscript{4} School of Data Science, Fudan University, Shanghai, China \\
  \textsuperscript{5} Institution for Big Data, Fudan University, Shanghai, China \\
  \texttt{\{ratie25, wbxiong24, cyjiang25\}@m.fudan.edu.cn,} \\
  \texttt{\{zhengyushi, xysu, wulibo, lihao\_lh\}@fudan.edu.cn}
}

\begin{document}

\maketitle

\begingroup
  \def\thefootnote{*}
  \footnotetext{Equal contribution. \quad \textsuperscript{$\dagger$}Corresponding authors.}
\endgroup

\begin{abstract}
Climate downscaling of General Circulation Models (GCMs) is an ill-posed inverse problem, as GCM outputs and reanalysis fields arise from different dynamical systems and lack fine-scale temporal pairing. Supervised methods thus rely on synthetic degradations or model-specific adaptation, limiting generalization across heterogeneous GCMs. Zero-shot diffusion posterior sampling (DPS) avoids paired supervision by combining a high-resolution climate prior with inference-time constraints from coarse GCM inputs, but faces consistency and efficiency challenges. We propose GCM-consistent Diffusion for Zero-shot Downscaling (GCD), which addresses these challenges with geographic conditioning, spectrally filtered measurement guidance, and distillation-accelerated Conjugate Gradient guidance. GCD reduces inference from 1,000 diffusion steps to 6 while maintaining consistency with the coarse GCM inputs. Across five heterogeneous GCMs, GCD achieves competitive 99th-percentile extreme-event accuracy and reconstructs plausible high-frequency tropical cyclone structures without model-specific retraining. Github: \url{https://github.com/shawnvonbohr/GCD-Climate-Downscaling.git}
\end{abstract}

\section{Introduction}
\label{Introduction}

Climate downscaling is a fundamental task~\cite{hewitson1996climate} that seeks to reconstruct plausible high-resolution atmospheric states from coarse-resolution (e.g., hundreds of kilometers) General Circulation Model (GCM)~\cite{meehl2000coupled} outputs, supporting downstream applications such as climate impact assessment~\cite{kates1987climate}.

Despite its importance, climate downscaling remains fundamentally challenging due to the mismatch between GCM simulations and reanalysis data. 
Specifically, GCMs are free-running physical simulations whose generated weather trajectories do not chronologically align with reanalysis datasets such as ERA5.
As a result, treating reanalysis data at the same timestamp as ground-truth labels introduces intrinsic supervision noise, making direct end-to-end training infeasible~\cite{li_generative_2025, tie2025generative, schmidt2025generative}.

\begin{figure}[t]
  \centering
  \includegraphics[width=0.8\textwidth]{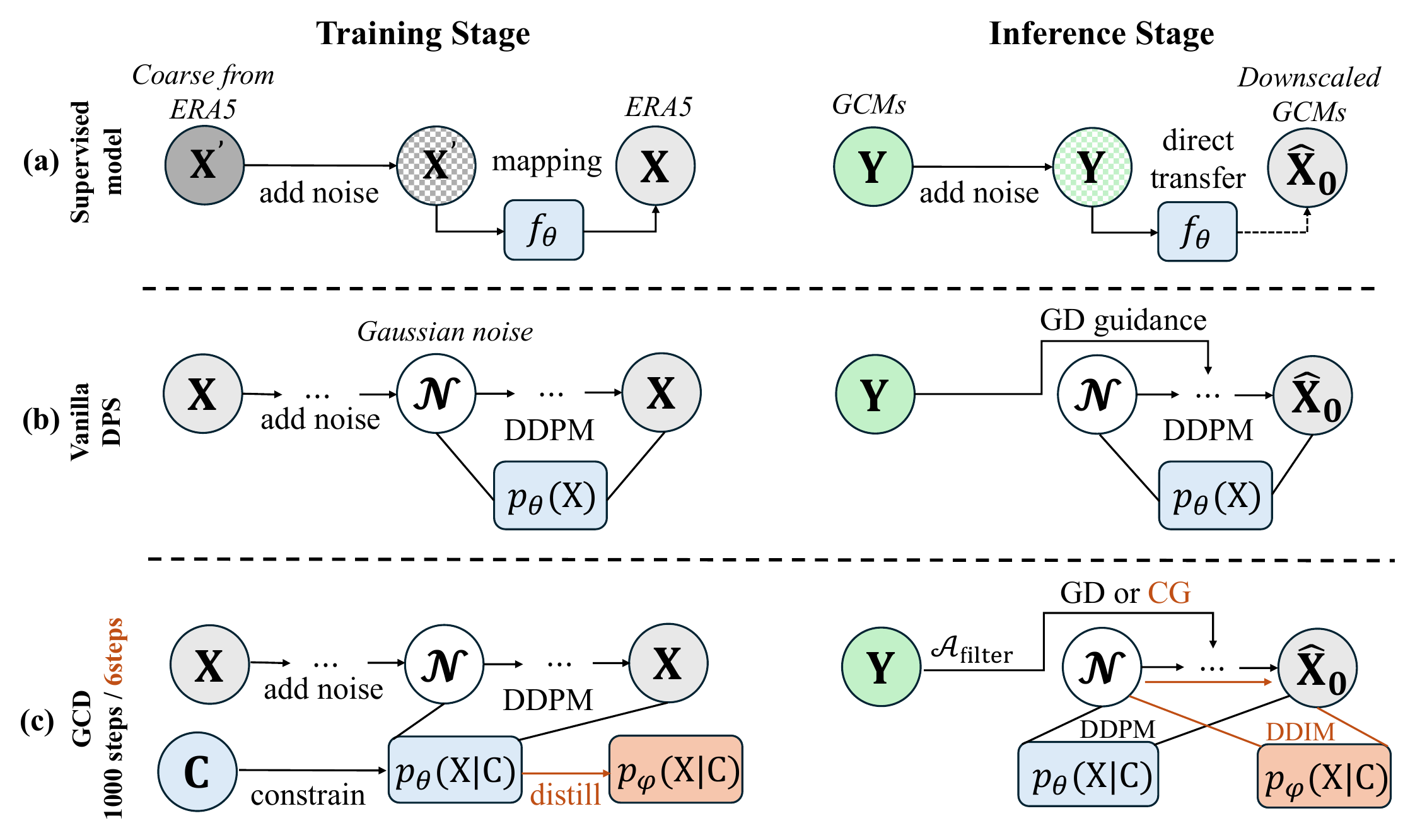}
  \caption{\small Schematic comparison of different downscaling paradigms during training and inference.}
  \label{fig1}
  \vspace{-0.4cm}
\end{figure}

Previous studies address the mismatch between GCMs and reanalysis data by training downscaling models on proxy low-resolution inputs constructed from reanalysis data. Specifically, high-resolution reanalysis fields are artificially coarsened to generate synthetic low-resolution inputs, while the original reanalysis fields are used as high-resolution targets. The learned mapping is then applied directly to GCM outputs at inference time~\cite{vandal2017deepsd,hess2025fast}, as shown in Fig.~\ref{fig1}(a).
However, the learned mapping implicitly depends on the specific degradation process used to generate the synthetic inputs, which does not faithfully reflect the diverse error characteristics of real GCMs. As a result, a strong coupling exists between the trained model and a particular GCM, limiting its ability to generalize across heterogeneous GCMs with different spectral biases.
This limitation becomes more pronounced in practical climate impact assessment, which typically relies on ensemble analysis over multiple heterogeneous GCMs, often ranging from 10 to 30 models~\cite{trisos2020projected,kotz2024retracted,mora2017global}.
Applying such approaches in this setting would require repeated retraining or adaptation for each GCM, resulting in poor scalability and substantial computational overhead.

Recent studies formulate GCM downscaling as an inverse problem and utilize diffusion posterior sampling (DPS)~\cite{chung2022diffusion,schmidt2025generative} to overcome this limitation. By learning an unconditional prior from higher-resolution reanalysis data and enforcing GCM consistency at inference time, DPS improves generalization across heterogeneous GCMs without model-specific retraining, as shown in Fig.~\ref{fig1}(b). 
However, directly applying vanilla DPS to global climate downscaling is bottlenecked by two critical challenges: consistency and efficiency. For consistency, the unconditional prior lacks geographic constraints and can produce physically implausible spatial patterns; meanwhile, standard measurement operators ignore spectral discrepancies between GCMs and the prior, degrading cross-model performance. For efficiency, the reverse diffusion process requires hundreds to thousands of function evaluations (NFEs), resulting in prohibitive inference costs. Beyond its application value, this setting poses a general machine learning challenge: zero-shot posterior sampling under domain shift, where a single pretrained diffusion prior must serve many heterogeneous, out-of-distribution measurement sources without retraining.

Motivated by the above challenges, we propose \textbf{GC}M-consistent Diffusion for Zero-shot \textbf{D}ownscaling across Heterogeneous GCMs (GCD), a zero-shot statistical downscaling framework for global GCMs. 
By eliminating the need for laborious model-specific retraining, GCD can be directly applied to diverse heterogeneous GCMs with substantially reduced computational costs. 
GCD targets cross-model consistency and efficient inference through three components: (1)~\emph{a geographically-constrained prior}, (2)~\emph{a filtering measurement operator}, and (3)~\emph{progressive distillation with conjugate gradient (CG) guidance}, as shown in Fig.~\ref{fig1}(c). 
Our contributions are outlined as follows:

\paragraph{-- A geographically-constrained climate prior to restore spatial consistency.} We integrate static geographic boundaries ($\mathbf{C}$, e.g., topography) into the unconditional prior. Elevation and land--sea information has long been used in statistical downscaling~\cite{vandal2017deepsd}; we inject it into the score network of a diffusion prior. This explicit conditioning reduces physically implausible patterns and anchors the generated structures to local geography.

\paragraph{--  A filtering measurement operator to resolve spectral inconsistency.} We replace the asymmetric downsampling operator in vanilla DPS with a filtering measurement operator that acts as a symmetric low-pass filter. By extracting and aligning only the reliable large-scale signals (e.g., 5$^\circ$ resolution) from both the estimated $\hat{\mathbf{X}}_0$ and GCM inputs, this operator substantially reduces the high-frequency spectral discrepancies between the ERA5 prior and heterogeneous models. Guidance thus constrains only scales at or above $5^\circ$; finer-scale structure is supplied by the geographically conditioned prior. 

\paragraph{-- A joint acceleration scheme for efficient posterior sampling.} We combine progressive distillation with CG guidance to accelerate both the prior and guidance processes. For the prior, we first identify an effective NFE for the DDIM sampler (25 steps in our ablation) that preserves generation quality relative to the 1,000-step baseline, and then distill a compact prior that operates at this reduced step budget. 
For guidance, we replace vanilla gradient descent with the conjugate gradient (CG) method, which more effectively solves the underlying inverse problem and compensates for the truncation errors introduced by few-step sampling. Crucially, these two components are designed to be complementary: the distilled prior reduces the number of required sampling steps, while CG guidance improves convergence under limited iterations. 
Together, they compress inference from 1,000 steps to 6 while maintaining GCM consistency and competitive generation quality on most GCMs.

\paragraph{-- Empirical evidence.} Experiments on five CMIP6 GCMs show that GCD (1,000 NFE) is competitive with or better than supervised baselines in 99th-percentile errors, and the 6-step variant remains competitive on four of five GCMs, with a degradation on IPSL, whose raw fields deviate strongly from the ERA5 prior. GCD also reconstructs plausible high-frequency structures of tropical cyclones (TCs) with spectra consistent with ERA5. Because GCMs are free-running, we evaluate distributional fidelity rather than event-wise correctness.

\section{Related Work}
\textbf{Unpaired Learning for Climate Downscaling.}
Supervised statistical downscaling trains on artificially coarsened reanalysis~\citep{vandal2017deepsd,hess2025fast}, but the learned mapping is tied to the degradation used in training. Since GCM simulations and real-world observations lack paired data, researchers have explored unpaired and zero-shot downscaling. Early approaches aligned domain distributions via Normalizing Flows \citep{groenke2020climalign} or CycleGANs \citep{li_generative_2025}. Recently, \citet{schmidt2025generative} formulated downscaling as a Bayesian inverse problem using vanilla Diffusion Posterior Sampling (DPS). However, applying vanilla DPS to global GCMs remains bottlenecked by physically implausible patterns from unconditional priors, spectral biases introduced by standard measurement operators, and prohibitive sampling costs. Our proposed GCD framework is designed to address these three limitations.

\textbf{Solving Inverse Problems via Diffusion Priors.}
Leveraging diffusion models as priors for inverse problems has evolved from exploiting operator linearity to handling general non-linear degradations. For linear tasks, \citet{kawar2022denoising} and \citet{wang2022zero} proposed DDRM and DDNM, respectively, which utilize SVD or null-space decomposition to analytically enforce data consistency. To extend applicability to general operators where such decomposition is infeasible, 
building upon the gradient guidance strategy in \citet{chung2022improving}, \citet{chung2022diffusion} introduced DPS, which approximates the likelihood score to guide the reverse process.
Addressing the high computational cost of such gradient-based guidance, \citet{zhu2023denoising} focused on acceleration strategies, while \citet{mardani2023variational} proposed variational formulations that optimize the evidence lower bound directly. Closest to our guidance design, DDS~\citep{chung2024decomposed} uses conjugate gradient (CG) to accelerate diffusion-based solvers for large-scale inverse problems. We build on the same idea but differ in setting and design: we couple CG with a spectral low-pass measurement operator and a distilled few-step student, targeting zero-shot downscaling across heterogeneous GCMs. WIND~\citep{aich2026wind} is a concurrent diffusion-based framework for zero-shot atmospheric inverse problems; we discuss it as related work and do not compare against it.

\textbf{Few-Step Generation.}
Progressive distillation~\citep{salimans2022progressive} and consistency models~\citep{song2023consistency} reduce sampling cost by learning large-stride transitions. We adopt progressive distillation rather than training a few-step model directly, for three reasons: (i) the student retains a discrete chain of intermediate noisy states, at which measurement guidance can be injected at inference time; (ii) teacher outputs allow explicit supervision with gradient and frequency losses to preserve high-frequency structure; and (iii) no per-GCM retraining is needed. Consistency models (CM, CM+noise) are included as baselines in Sec.~\ref{sec:Experiments}.

\section{Methodology}

\begin{figure*}[!t]
  \vskip -0.1in
  \begin{center}
    \centerline{\includegraphics[width=\textwidth]{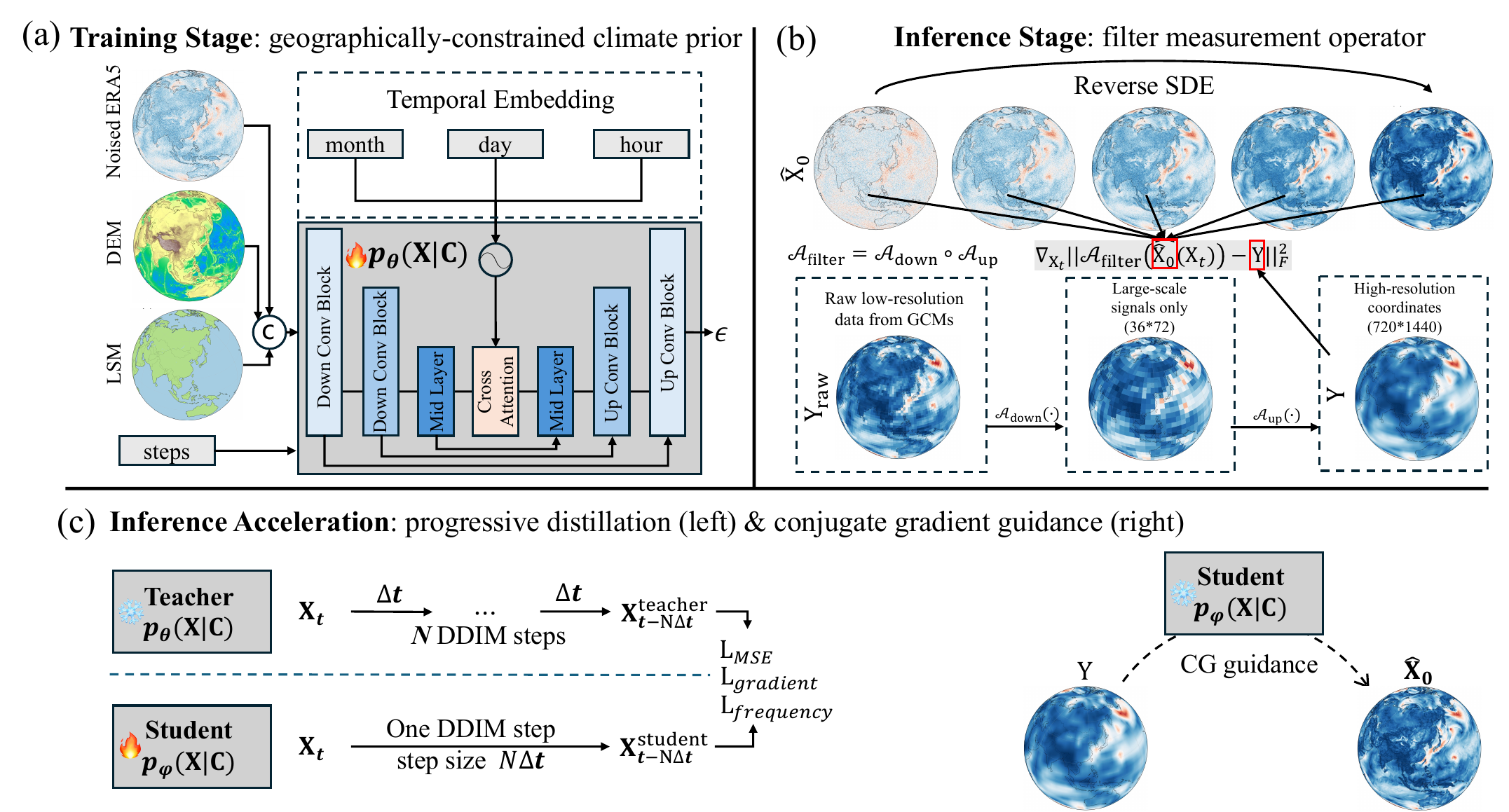}}
    \caption{\small
Overview of the GCD framework. 
(a) Training Stage: a geographically-constrained climate diffusion prior conditioned on static terrain and temporal embeddings.
(b) Inference Stage: a filter measurement operator is applied during reverse diffusion sampling to reduce the spectral bias between the prior and GCMs.
(c) Inference Acceleration: we use distillation and conjugate gradient guidance to compress the teacher into a fast student model.
    }
    \label{fig2}
  \end{center}
  \vspace{-1cm}
\end{figure*}

We present the GCD framework, with its overall training and inference pipelines illustrated in Fig.~\ref{fig2}. We first formulate the downscaling task as an inverse problem (Section~\ref{subsec:Problem Formulation}). Next, we detail the geographically-constrained climate prior and the filter measurement operator used to guide the generation process (Section~\ref{subsec:prior_and_measurement}). Finally, we introduce a joint acceleration scheme combining distillation and conjugate gradient guidance to improve inference efficiency (Section~\ref{subsec: Inference Accelerate}).

\subsection{Problem Formulation}
\label{subsec:Problem Formulation}

\textbf{Climate Downscaling as Inverse Problem.}
Let $\mathbf{Y} \in \mathbb{R}^{h \times w \times C}$ denote a coarse-resolution GCM output, where $h$ and $w$ are the spatial grid dimensions (latitude and longitude; they vary across GCMs, see Appendix~\ref{Details of Datasets}), and $C$ is the number of physical variables (wind components, pressure, and geopotential; $C=5$ in our experiments).
The corresponding high-resolution state is $\mathbf{X} \in \mathbb{R}^{H \times W \times C}$ on a finer grid, with $H=720$, $W=1440$ (ERA5 at $0.25^\circ$), $h \ll H$ and $w \ll W$.
Climate downscaling aims to reconstruct plausible fine-scale states from coarse-resolution simulations.
Although $\mathbf{X}$ and $\mathbf{Y}$ are three-dimensional tensors, we represent them as two-dimensional matrices for notational simplicity in the following text, without loss of generality.
The task can be formulated as an ill-posed inverse problem, where the coarse-resolution field $\mathbf{Y}$ is degraded from the high-resolution state $\mathbf{X}$ through a non-invertible measurement operator:

{\small
\begin{equation}
    \mathbf{Y} = \mathcal{A}(\mathbf{X}) + \mathbf{N},
    \label{eq:degradation_model}
\end{equation}
}

where $\mathcal{A}(\cdot): \mathbb{R}^{H \times W} \mapsto \mathbb{R}^{h \times w}$ models spatial aggregation and resolution transform, $\mathbf{N} \sim \mathcal{MN}_{h \times w}(\mathbf{0}, \boldsymbol{\Sigma}_h, \boldsymbol{\Sigma}_w)$ is measurement noise following a matrix variate Gaussian distribution, and $\boldsymbol{\Sigma}_h$ and $\boldsymbol{\Sigma}_w$ are the row (latitudinal) and column (longitudinal) covariance matrices, respectively.

The objective is to learn a model $f_\theta$ that, given $\mathbf{Y}$, produces a high-resolution reconstruction $\hat{\mathbf{X}}$ consistent with the large-scale content of $\mathbf{Y}$ and statistically plausible at fine scales, rather than merely increasing resolution through interpolation.

\textbf{Vanilla Diffusion Posterior Sampling in Climate Downscaling.}
We seek to reconstruct a plausible high-resolution state $\mathbf{X}$ from a coarse-resolution GCM output $\mathbf{Y}$ without access to paired coarse- and high-resolution samples. The ill-posed nature of the problem motivates a probabilistic formulation, in which the solution is represented by the posterior distribution $p(\mathbf{X} | \mathbf{Y})$ rather than a point estimate. By Bayes' rule,

{
\small
\begin{equation}
    p(\mathbf{X} | \mathbf{Y}) \propto 
    \underbrace{p(\mathbf{Y} | \mathbf{X})}_{\text{Data Consistency}} 
    \cdot 
    \underbrace{p(\mathbf{X})}_{\text{Climate Prior}},
    \label{eq:bayes_rule}
\end{equation}
}

where $p(\mathbf{X})$ is a prior over physically and statistically plausible high-resolution states, and the likelihood $p(\mathbf{Y} | \mathbf{X})$ encodes the consistency between the reconstructed field and the observed coarse-resolution GCM output. In this zero-shot setting, the prior is trained on ERA5 alone, and the GCM output enters only at inference time, without paired supervision.

\subsection{Geographically-Constrained Climate Prior \& Filter Measurement Operator}
\label{subsec:prior_and_measurement}

Unlike vanilla DPS formulations that employ an unconditional diffusion prior, we parameterize the prior using a conditional DDPM~\cite{NEURIPS2020_4c5bcfec}. To incorporate geographical information, we condition the reverse process on a context $\mathbf{C}$. This context integrates static spatial boundaries, specifically the land--sea mask (LSM) and digital elevation model (DEM), via channel-wise concatenation, in line with the use of elevation information in statistical downscaling~\cite{vandal2017deepsd}. It also includes cyclic temporal information (month, day, and hour) encoded via sine--cosine embeddings and cross-attention (see Fig.~\ref{fig2}(a)). Given high-resolution reanalysis data $\mathbf{X}_0$, the network $\boldsymbol{\epsilon}_\theta(\mathbf{X}_t, t, \mathbf{C})$ is trained to approximate the added Gaussian noise (details in Appendix~\ref{SI: Supplementary Theory 3.1}).

During inference, vanilla DPS relies on a downsampling operator $\mathcal{A}_{\text{down}}$ to align the estimated state $\hat{\mathbf{X}}_0$ with coarse GCM outputs~\cite{schmidt2025generative}. Inspired by~\citet{hess2025fast}, we observe that this operator overlooks the spectral bias between the ERA5-trained prior and heterogeneous GCMs. These discrepancies in high-frequency spatial signals degrade generation performance. Frequency-domain filtering, such as the Fast Fourier Transform (FFT), is a standard way to remove high frequencies, but hard frequency cutoffs on meteorological fields introduce ringing artifacts (the Gibbs phenomenon).

To avoid frequency-domain filtering, we propose a spatial-domain Filter Measurement Operator, $\mathcal{A}_{\text{filter}} = \mathcal{A}_{\text{up}} \circ \mathcal{A}_{\text{down}}$. By applying area-weighted downsampling to a $5^\circ$ resolution ($36 \times 72$ grid) before upsampling, $\mathcal{A}_{\text{filter}}$ acts as a spatial low-pass filter. It reduces high-frequency discrepancies and avoids the ringing of hard spectral cutoffs, providing large-scale guidance for the generative process. The same operator is applied to the generated field and to the GCM input (after regridding to the common $720\times1440$ grid), so both are compared at the same $5^\circ$ scale. The inference details and algorithms of GCD are provided in Appendix~\ref{Inference Algorithm of ZSSD}.

\textbf{What is constrained.} Guidance constrains only scales at or above $5^\circ$. Structure below $5^\circ$ is supplied by the ERA5-trained, geographically conditioned prior. We therefore evaluate generated fields distributionally and do not claim event-level correctness (Sec.~\ref{sec:Experiments}).

\subsection{Distillation \& Conjugate Gradient Guidance}
\label{subsec: Inference Accelerate}

To accelerate sampling, we distill a fast student model from a DDIM teacher. To limit distillation cost, we first identify an effective DDIM step budget that balances speed and generation quality (25 steps in our ablation, Sec.~\ref{sec:Experiments}). The model at this ``sweet spot'' serves as the teacher. The student is trained in a single distillation stage to reproduce the teacher's cumulative transition over $N$ consecutive teacher steps (stride $\Delta t$) in one step of stride $N\Delta t$, which compresses the 25-step teacher into a 6-step student on a non-uniform time schedule. Since distillation with a plain mean squared error ($\mathcal{L}_{\text{MSE}}$) on the intermediate denoised states $\mathbf{X}_t$ tends to yield over-smoothed results, we use a joint loss:

{
\small
\begin{equation}
 \mathcal{L}_{\text{total}} = \mathcal{L}_{\text{MSE}} + \lambda_{\text{grad}} \mathcal{L}_{\text{grad}} + \lambda_{\text{freq}} \mathcal{L}_{\text{freq}}.  
    \label{eq: joint loss function}
\end{equation}
}

The term $\mathcal{L}_{\text{MSE}}$ encourages global pixel consistency between the student states $\mathbf{X}_{t-N\Delta t}^{\text{student}}$ and the teacher states $\mathbf{X}_{t-N\Delta t}^{\text{teacher}}$:

{
\small
\begin{equation}
 \mathcal{L}_{\text{MSE}} = \| \mathbf{X}_{t-N\Delta t}^{\text{student}} - \mathbf{X}_{t-N\Delta t}^{\text{teacher}} \|_F^2 .
    \label{eq: loss of Mean Squared Error}
\end{equation}
}

\begin{figure}[t!]
  \centering
  \includegraphics[width=1\textwidth]{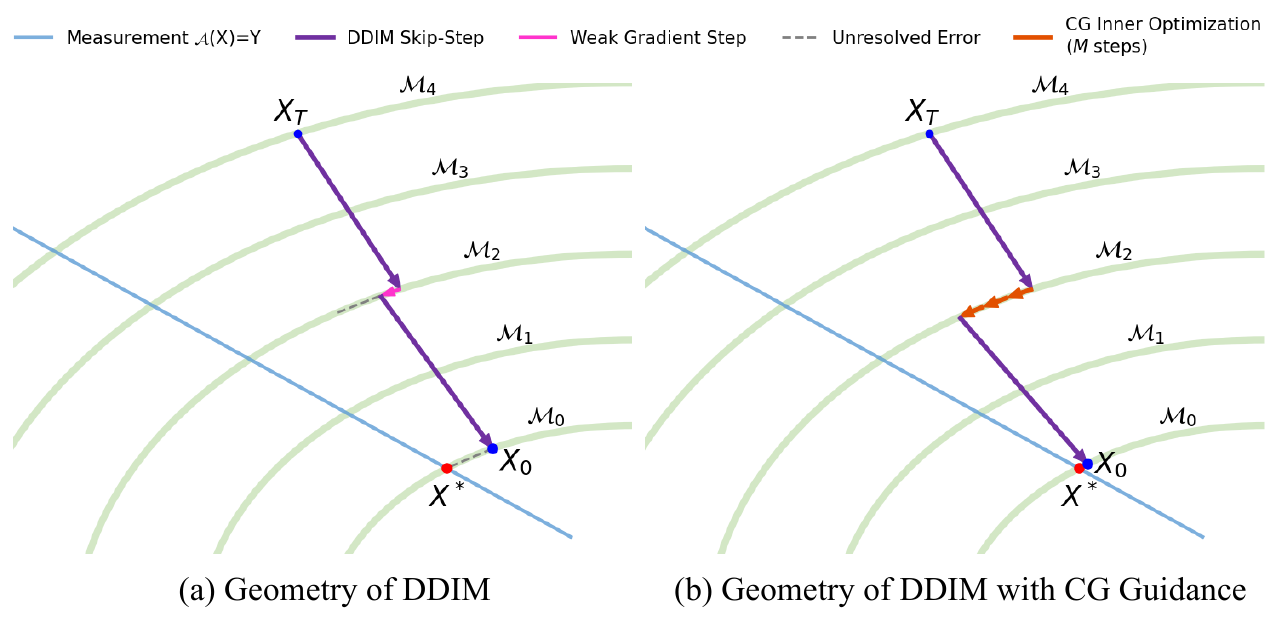}
  \caption{\small Schematic illustration of the geometries of the two guidance processes (not measured data).}
  \label{fig:cg_guidance}
  \vspace{-0.5cm}
\end{figure}

To preserve edge structures, the spatial gradient loss $\mathcal{L}_{\text{grad}}$ penalizes differences in local spatial variations of the predicted clean state $\hat{\mathbf{X}}_0$:

{
\small
\begin{equation}
 \mathcal{L}_{\text{grad}} = \| \nabla \hat{\mathbf{X}}_0(\mathbf{X}_{t-N\Delta t}^{\text{student}}) - \nabla \hat{\mathbf{X}}_0(\mathbf{X}_{t-N\Delta t}^{\text{teacher}}) \|_F^2 ,
    \label{eq: loss of grad}
\end{equation}
}

where $\nabla$ denotes the spatial gradient operator (e.g., Sobel filter). The frequency loss $\mathcal{L}_{\text{freq}}$ is computed in the Fourier domain to counter the loss of high-frequency texture:

{
\small
\begin{equation}
 \mathcal{L}_{\text{freq}} = \| \mathcal{F}(\hat{\mathbf{X}}_0(\mathbf{X}_{t-N\Delta t}^{\text{student}})) - \mathcal{F}(\hat{\mathbf{X}}_0(\mathbf{X}_{t-N\Delta t}^{\text{teacher}})) \|_F^2 ,
    \label{eq: loss of freq}
\end{equation}
}

where $\mathcal{F}$ denotes the FFT. Ablations (Sec.~\ref{sec:Experiments}) show that adding $\mathcal{L}_{\text{grad}}$ and $\mathcal{L}_{\text{freq}}$ lowers the 6-step error compared with $\mathcal{L}_{\text{MSE}}$ alone.

While distillation reduces the number of denoising steps, consistency between the generated results and the observation must still be enforced. For notational simplicity, let $\mathbf{X}$ denote the intermediate denoised state at the current step and $\hat{\mathbf{X}}_0(\mathbf{X})$ the associated predicted clean state. Vanilla DPS updates $\mathbf{X}$ along the negative gradient of $\mathcal{J}(\mathbf{X}) = \|\mathcal{A}(\hat{\mathbf{X}}_0(\mathbf{X})) - \mathbf{Y}\|_F^2$ with gradient descent (GD). In the few-step regime, a single GD update per step corrects the deviations of large-stride prior transitions less accurately, and the residual errors accumulate across steps (Fig.~\ref{fig:cg_guidance}(a); ablation in Sec.~\ref{sec:Experiments}).

We therefore replace GD with the conjugate gradient (CG) method, applied to a regularized least-squares problem. Rather than relying on a local gradient step ($\mathbf{X} \leftarrow \mathbf{X} - \eta \nabla \mathcal{J}(\mathbf{X})$), we solve an inner-loop problem that strengthens observation consistency while staying close to the prior estimate:

{
\small
\begin{equation}
\min_\mathbf{X} \|\mathcal{A}(\mathbf{X}) - \mathbf{Y}\|_F^2 + \lambda \|\mathbf{X} - \hat{\mathbf{X}}_0\|_F^2 ,
    \label{eq: optimization objective of CG}
\end{equation}
}

where $\lambda>0$ controls how strongly the solution is anchored to $\hat{\mathbf{X}}_0$. This is a Tikhonov regularization centered on $\hat{\mathbf{X}}_0$. Setting the gradient to zero gives the normal equation

{
\small
\begin{equation}
(\mathcal{A}^T\mathcal{A} + \lambda \mathbf{I})\mathbf{X} = \mathcal{A}^T\mathbf{Y} + \lambda \hat{\mathbf{X}}_0 .
    \label{eq: normal equation}
\end{equation}
}

For $\lambda>0$ the matrix $\mathcal{A}^T\mathcal{A}+\lambda\mathbf{I}$ is symmetric positive definite, so CG applies, and its condition number is bounded by $(\|\mathcal{A}\|_2^2+\lambda)/\lambda$. We run $M=5$ CG iterations with $\lambda=0.1$. Each iteration requires one application of $\mathcal{A}$ and one of $\mathcal{A}^T$; since $\mathcal{A}_{\text{filter}}$ is a sparse, local linear operator on $\mathbf{X}\in\mathbb{R}^{720\times1440\times5}$, the cost per iteration is $O(HWC)$.

The CG solver relies on the linearity of $\mathcal{A}$; nonlinear operators would require other solvers or standard DPS-style gradient guidance. The overall procedure is summarized in Fig.~\ref{fig2}(c), with details in Appendix~\ref{Inference Algorithm of Conjugate Gradient Acceleration}.

\section{Experiments}
\label{sec:Experiments}

\subsection{Experimental Setup}
\textbf{Datasets.} \underline{\emph{GCMs Simulations.}} To evaluate the zero-shot generalization capability of our framework, we use 6-hourly outputs from the historical experiment of five different {GCMs models in Coupled Model Intercomparison Project Phase 6 (CMIP6)} \cite{eyring2016overview}, including IPSL-CM6A-LR \cite{boucher2020presentation}, AWI-ESM-1-1-LR \cite{semmler2020simulations}, MIROC6 \cite{tatebe2019description}, MPI-ESM1-2-LR and MPI-ESM1-2-HR \cite{mauritsen2019developments}. \underline{\emph{ERA5.}} We utilize the {ERA5} reanalysis dataset \cite{hersbach2023era5} as our high-resolution ground truth, using its native $0.25^\circ$ spatial resolution ($720 \times 1440$ grid points) and subsampling it to 6-hour intervals. The prior trained on ERA5 (1980–1999), validate on 2000, and test on 2001–2009 using both ERA5 and GCMs. GCMs data are not used in training or validation. More details about variables are in Appendix~\ref{Details of Datasets}

\textbf{Baselines.} 
We benchmark GCD against three distinct paradigms. First, traditional interpolations (Bilinear, Cubic). Second, supervised methods, which we evaluate in a direct-transfer setting to assess their cross-GCM generalization. This includes standard models (SwinIR, ESRGAN, CorrDiff, CM) that typically require model-specific retraining, as well as a domain-adapted variant (CM+noise) that injects artificial noise during training to mitigate domain gaps. Third, zero-shot solvers (DDRM, vanilla DPS) that bypass paired training entirely. Details are provided in Appendix~\ref{Settings Details of Baselines}.

\textbf{Implementation Details.} 
Our prior is parameterized by a time- and condition-aware 2D U-Net. All models are trained using the AdamW optimizer. Comprehensive architectural details and training hyperparameters are detailed in Appendix~\ref{Settings Details of ZSSD}.

\textbf{Evaluation Methods.} Following recent established practices for evaluating unaligned climate fields~\cite{schmidt2025generative}, we evaluate distributional consistency and extreme-state characteristics by computing the latitude-weighted Mean Absolute Error (MAE) and Root Mean Square Error (RMSE) of the 99th percentile for each variable (formulas in Appendix~\ref{Supplementary Evaluation Methods}). To reduce the impact of randomness, we calculate the average over 5 ensemble members.

\subsection{Main Results}

We evaluate GCD's zero-shot capabilities directly across five heterogeneous GCMs through two critical dimensions. First, we benchmark overall performance using 99th percentile errors to assess extreme climate simulations, alongside a comparative analysis of inference efficiency and training costs. Second, we validate physical fidelity across scales by visualizing localized TCs structures and analyzing the global annual mean Power Spectral Density (PSD) to confirm the restoration of high-frequency atmospheric energy.

\begin{table*}[t]
\vskip -0.2in
  \caption{\small Quantitative comparison of downscaling methods on different GCMs. The first five columns report the spatial MAE/RMSE averaged over the test period (2001-2009) for each respective GCM. Mean Inf. Time denotes the average inference time per sample across all five GCMs. Per GPU Hours reports the computational cost on a single GPU in training stage. Notably, the `X' multiplier indicates that the cost must be multiplied by the number of target GCMs (e.g., requiring separate training for each specific GCM). Best results are highlighted in \textbf{bold} and \underline{underline} denotes the second-best.}
  
  \label{tab:main_results}
  \begin{center}
    \begin{footnotesize}
      \begin{sc}
        \setlength{\tabcolsep}{4pt}
        \resizebox{\textwidth}{!}{%
        \begin{tabular}{c ccccc | cc}
          \toprule
          Method & IPSL & MIROC6 & AWI & MPI-LR & MPI-HR & \makecell[c]{\scriptsize Mean Inf. Time \\ \scriptsize per Sample (s)} & \makecell[c]{\scriptsize per \\ \scriptsize GPU Hours (h)} \\
          \midrule
          Bilinear & 1.06 / 1.78 & 1.84 / 2.43 & 1.50 / 2.19 & 1.48 / 2.12 & 1.32 / 1.90 & $\mathbf{1.80 \times 10^{-5}}$ & -- \\
          Cubic    & 1.03 / 1.79     & 1.82 / 2.42     & 1.50 / 2.18     & 1.47 / 2.12     & 1.30 / 1.87     & \underline{$1.37 \times 10^{-4}$}   & -- \\
          BCSD~\cite{wood2002long}     & 0.98 / 1.67 & 1.72 / 2.33 & 1.36 / 2.07 & 1.35 / 1.98 & 1.24 / 1.76 & 0.27 & -- \\
          \midrule
          SwinIR~\cite{liang2021swinir}   & 1.01 / 1.67     & 1.52 / 2.05     & 1.37 / 2.02     & 1.35 / 1.95     & 1.18 / 1.68     & 0.25   & 11X \\
          ESRGAN~\cite{wang2018esrgan}   & 1.00 / 1.66     & 1.50 / 2.03     & 1.36 / 2.00     & 1.34 / 1.93     & 1.17 / 1.67     & 0.04   & 22X \\
          \makecell[c]{CorrDiff~\cite{mardani2025residual}} & 0.98 / 1.62     & 1.45 / 1.95     & 1.33 / 1.96     & 1.31 / 1.89     & 1.14 / 1.62     & 2.33   & 35X \\
          \makecell[c]{CM~\cite{song2023consistencymodels}}       & 0.97 / 1.61     & 1.46 / 1.96     & 1.34 / 1.97     & 1.30 / 1.88     & 1.15 / 1.63     & 0.16   & 33X \\
          \midrule
          \makecell[c]{CM+noise~\cite{hess2025fast}} & \underline{0.93 / 1.52}     & 1.32 / 1.78     & 1.30 / 1.92     & 1.28 / 1.85     & 1.10 / \underline{1.56}     & 0.16   & 33 \\
          \midrule
          \makecell[c]{DDRM~\cite{kawar2022denoising}}     & 1.04 / 1.69 & 1.84 / 2.45 & 1.53 / 2.20 & 1.43 / 2.10 & 1.33 / 2.02 & 3.92 & \textbf{12.5} \\
          \makecell[c]{DPS~\cite{chung2022diffusion}}      & 1.03 / 1.67 & 1.84 / 2.40 & 1.48 / 2.13 & 1.40 / 2.08 & 1.32 / 1.89 & 588.38 & \textbf{12.5} \\
          \makecell[c]{GCD (1000 NFE)}       & \textbf{0.87 / 1.32} & \underline{1.08} / \textbf{1.42} & \underline{1.28 / 1.89} & \underline{1.24 / 1.81} & \textbf{1.05 / 1.49} & 589.62 & \textbf{12.5} \\
          \makecell[c]{GCD (6 NFE)} & 1.49 / 1.95 & \textbf{1.05} / \underline{1.50} & \textbf{1.14 / 1.71} & \textbf{1.07 / 1.63} & \underline{1.08} / 1.59 & 4.38 & \underline{24.5} \\
          \bottomrule
        \end{tabular}%
        }
      \end{sc}
    \end{footnotesize}
  \end{center}
\end{table*}

\textbf{Overall Performance.} We evaluate GCD across five heterogeneous GCMs (Tab.~\ref{tab:main_results}), focusing on the trade-offs among accuracy, training costs, and inference efficiency. First, in terms of accuracy, the full-trajectory GCD (1000 NFE) demonstrates robust zero-shot generalization. It not only outperforms traditional zero-shot baselines like Vanilla DPS but also matches or exceeds supervised methods across most metrics. For instance, on the MIROC6 dataset, GCD (1000 NFE) achieves an MAE of 1.08, which is lower than the supervised models CorrDiff (1.45) and CM+NOISE (1.32).

Second, regarding scalability, GCD avoids the model-specific retraining limitations common in supervised approaches. Because GCMs and ERA5 lack chronologically paired data, standard supervised models (e.g., CorrDiff and SwinIR) typically learn a mapping from artificially degraded ERA5 to high-resolution ERA5. To ensure reliable inference, this artificial degradation must align with the specific characteristics of the target GCM. Consequently, the significant domain shifts between different GCMs often necessitate retraining for each new model. As shown in Tab.~\ref{tab:main_results}, this results in a linear increase in computational cost (denoted by the `X' multiplier; e.g., CorrDiff requires 35 GPU hours per new GCM). While methods like CM+NOISE attempt to bypass retraining by injecting random noise to simulate domain shifts, simple statistical noise struggles to encapsulate complex physical discrepancies, which limits their generalization accuracy. In contrast, GCD requires only a single training phase on ERA5 data (12.5 h for the full-trajectory base model, and 24.5 h in total for the distilled 6-NFE version), and can be directly applied to any GCM in a zero-shot manner.

Finally, regarding inference efficiency, GCD (6 NFE) provides a practical solution that balances speed and accuracy. Standard zero-shot diffusion methods (e.g., Vanilla DPS) require nearly 588 seconds per sample due to heavy gradient computations. Through our joint acceleration strategy, GCD (6 NFE) reduces this inference time to 4.38 seconds. While this extreme step compression introduces higher variance on certain distributions, such as the IPSL dataset where the MAE increases to 1.49, GCD (6 NFE) maintains stable zero-shot accuracy on the remaining four GCMs (MIROC6, AWI, MPI-LR, and MPI-HR). In some cases, it even outperforms supervised baselines (e.g., achieving an MAE of 1.05 on MIROC6). Overall, although slightly slower than the deterministic CorrDiff (2.33 seconds), GCD (6 NFE) strikes a reasonable balance between eliminating retraining costs, improving inference efficiency, and maintaining generation quality.

\begin{figure*}[!t]
  \begin{center}
    \centerline{\includegraphics[width=\textwidth]{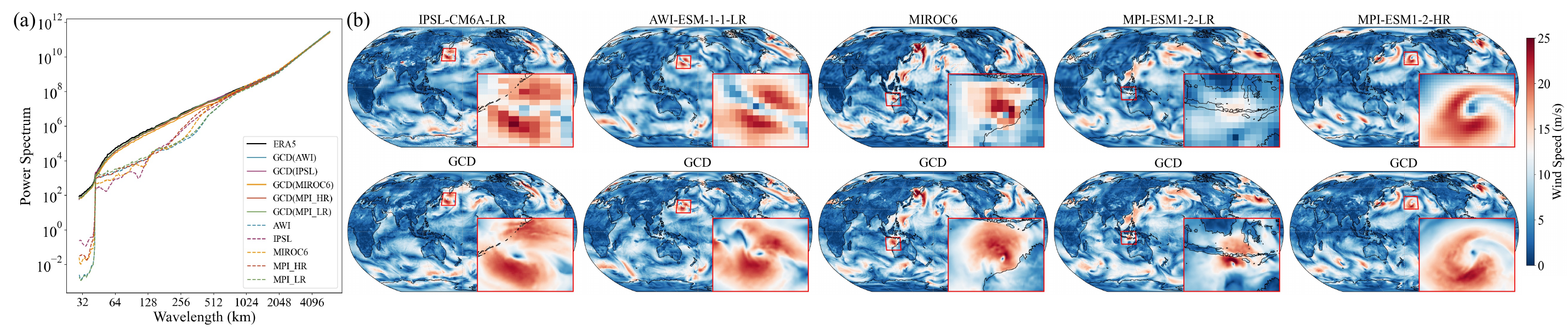}}
    \caption{ \small
    Spectral and spatial demonstration of GCD (1000 NFE) capabilities. 
    \textbf{(a)} Annual mean power spectra of 10m Wind Speed (WS10m).
    \textbf{(b)} Visual comparison of WS10m and TCs. The \textbf{top row} displays raw outputs from various GCMs. The \textbf{bottom row} shows the corresponding high-resolution fields reconstructed by our GCD framework, revealing detailed vortex structures. Columns correspond to samples from (left to right): 2001-02-24 00:00, 2002-01-26 00:00, 2001-01-01 00:00, 2001-01-03 00:00, and 2001-03-16 00:00.
    }
    \label{fig:tc_vis}
  \end{center}
  \vspace{-0.7cm}
\end{figure*}

\begin{figure*}[!t]
  \begin{center}
    \centerline{\includegraphics[width=\textwidth]{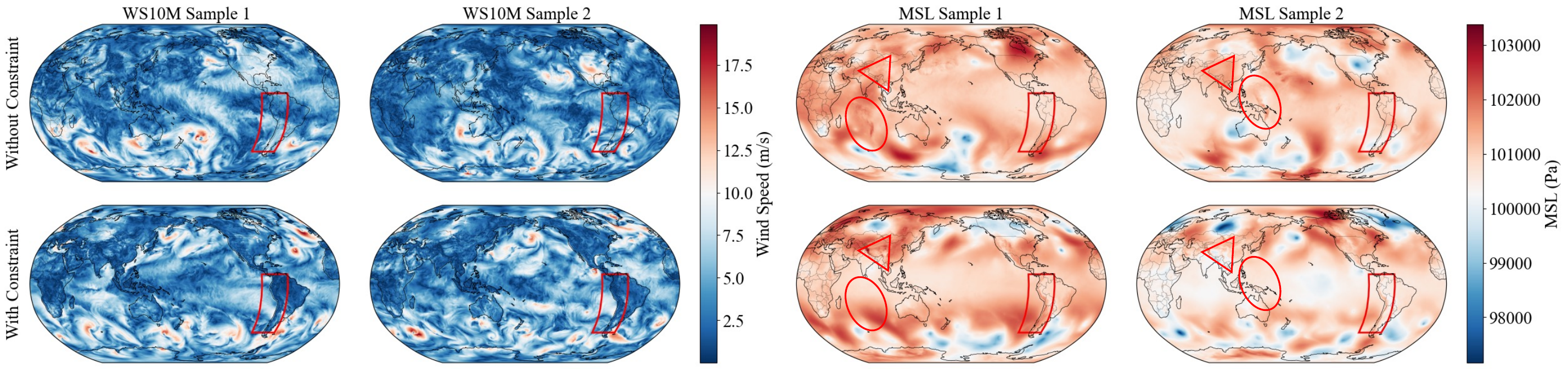}}
    \caption{\small
Impact of geographical constraints on generated climate states. The Top Row (without constraints) fails to capture topography-driven phenomena, lacking both the terrain blocking effect over the Andes (red box) and the expected high-pressure system over the Tibetan Plateau (red triangle), while also exhibiting spurious high-pressure artifacts (red ellipses). In contrast, the Bottom Row (with constraints) produces more physically plausible states, showing correct topography-induced wind deceleration, restoring missing pressure features, and eliminating unphysical anomalies.
}
    \label{ablation1.1}
  \end{center}
  \vspace{-1cm}
\end{figure*}

\textbf{TC Simulation.} Fig. \ref{fig:tc_vis} evaluates the zero-shot capabilities of GCD on TCs. As mesoscale weather systems, TCs exhibit significant energy concentrated at finer spatial scales, at scales of tens of kilometers. Fig. \ref{fig:tc_vis}(a) shows that raw GCMs experience substantial energy loss at these high-frequency bands. GCD effectively restores this missing energy, producing PSD curves that closely match the ERA5 reference at these critical scales. This spectral recovery directly corresponds to the spatial visualizations in Fig. \ref{fig:tc_vis}(b). By restoring the high-frequency energy, GCD reconstructs coherent vortex structures and detailed textures that are otherwise absent in the diffuse coarse inputs. Additional variables are provided in Appendix \ref{Supplementary Results in main results}.

\begin{wrapfigure}{r}{0.6\textwidth} 
  \vspace{-0.1in} 
  \begin{center}
    \includegraphics[width=\linewidth]{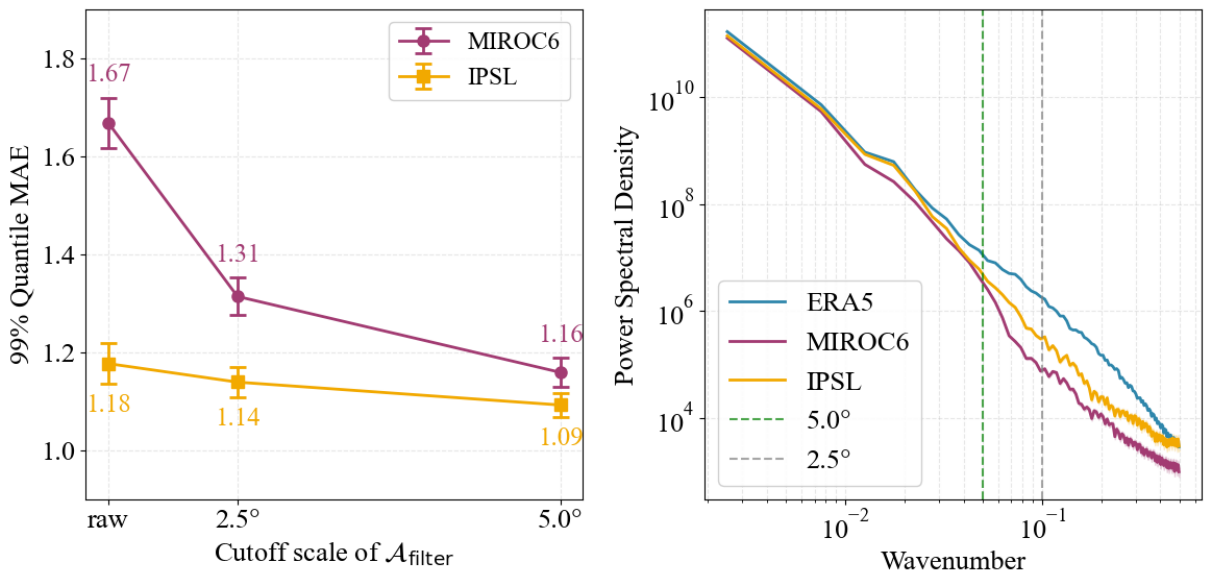} 
  \end{center}
  \vspace{-0.1in} 
  \caption{\small
Impact of cutoff scale of the filter measurement operator ($ \mathcal{A}_{\text{filter}} $). All evaluations are based on data from the test year 2001. (a) The 99\% quantile MAE evaluated at different cutoff scales: raw (representing the native resolution of the GCMs), 2.5°, and 5.0°. (b) PSD comparison among ERA5, MIROC6, and IPSL. The vertical dashed lines indicate the wavenumbers corresponding to the 5.0° and 2.5° scales.
  }
  \label{ablation3.1}
  \vspace{-0.1in} 
\end{wrapfigure}

\subsection{Ablation Study}

\textbf{Impact of Boundary Conditions on the Prior.} To validate the design of our geographically-constrained prior, we conduct an ablation study focusing on the boundary conditions. Fig. \ref{ablation1.1} compares random samples generated by the prior with and without geographical conditioning. Without these conditions, the generated wind fields are overly smooth and lack spatial variations associated with terrain features. Specifically, the flow patterns fail to reflect sharp transitions at coastlines or topographically induced constraints.

By contrast, conditioning on the Digital Elevation Model (DEM) and Land Sea Mask (LSM) yields structures that closely align with local geography. For example, along the west coast of South America, the conditioned model captures sharp gradients at the coastal boundaries and reflects the topographical influence of the Andes. These observations indicate that geographical conditioning is essential for preventing overly smoothed outputs and generating terrain-aware flow structures. A detailed breakdown of the individual contributions of DEM and LSM is provided in Appendix~\ref{Further Ablation Study About Prior}.

\textbf{Impact of Filter Measurment Operator.} Increasing the cutoff scale of the filter measurement operator ($ \mathcal{A}_{\text{filter}} $) effectively reduces the downscaling error. We evaluate this relationship using the 99\% quantile MAE on the 2001 test set. Fig.~\ref{ablation3.1}(a) compares the performance across three scales: the native GCM resolution (raw), 2.5°, and 5.0°. For both MIROC6 and IPSL, the error consistently decreases as the cutoff scale becomes larger. The lowest MAE is achieved at the 5.0° scale. This trend indicates that removing small-scale spatial information from the input variables improves the final prediction accuracy.

This performance improvement stems from mitigating the high-frequency spectral discrepancies between the GCMs and the ERA5 reference. Fig.~\ref{ablation3.1}(b) illustrates the PSD of the datasets. At higher wavenumbers corresponding to the raw resolutions, the GCM spectra deviate significantly from the ERA5 distribution. Applying a 5.0° low-pass filter successfully removes these mismatched high-frequency components. Consequently, the aligned large-scale features serve solely as inference-time guidance, preventing the downscaling process from being misled by inaccurate local inputs.

Additionally, unlike spatial downsampling measurement operator that suffers from vanishing gradients at the 5.0° scale, this filter operator maintains stable guidance gradients throughout the generative process (see Appendix~\ref{Further Ablation Study About Measurement Operator} for detailed gradient analysis).

\begin{figure}[!htbp]
    \vspace{-0.5cm}
    \centering
    \begin{minipage}[c]{0.43\textwidth}
        \centering
        \includegraphics[width=\linewidth]{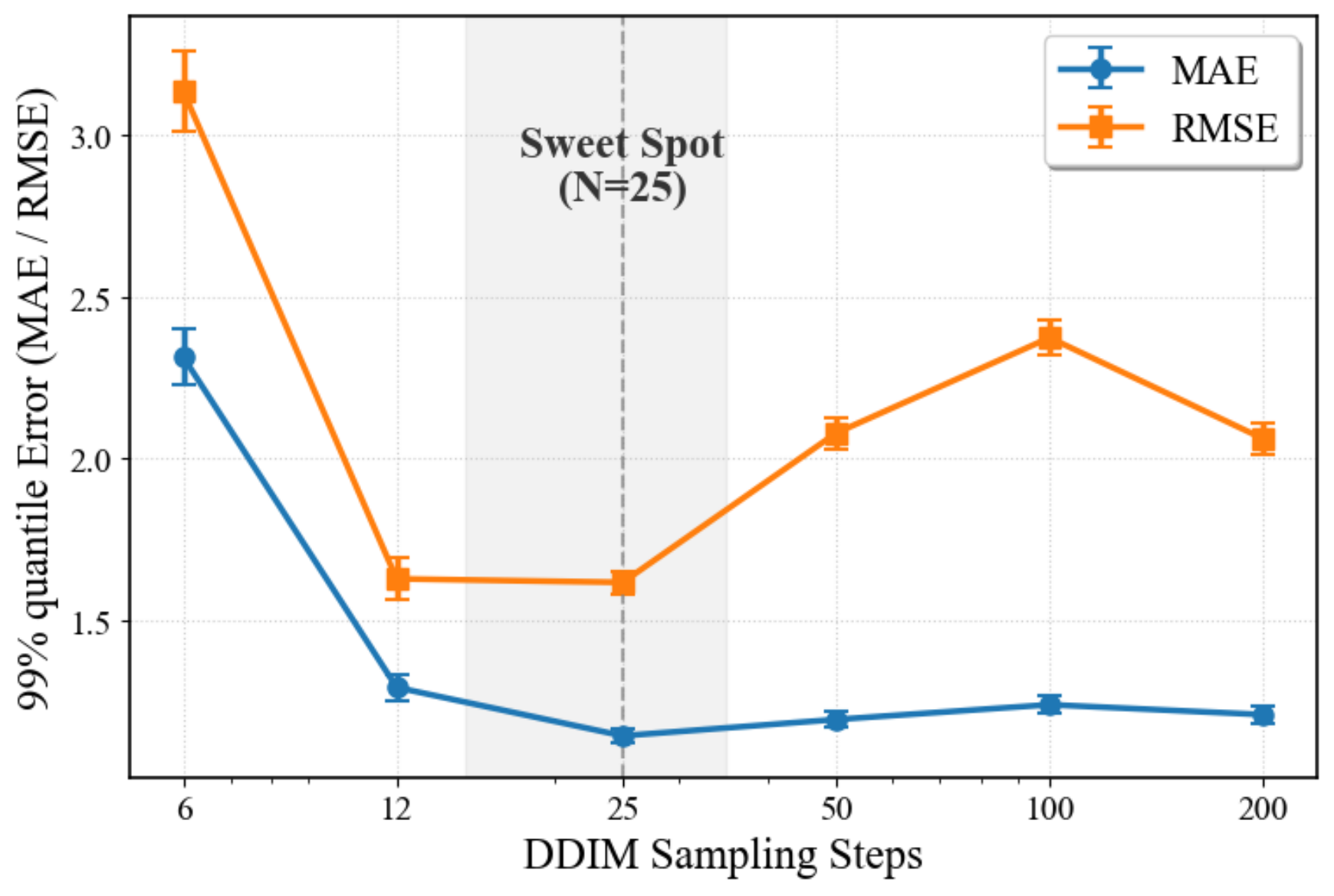}
    \end{minipage}
    \hfill
    \begin{minipage}[c]{0.55\textwidth}
        \centering
        \resizebox{\linewidth}{!}{
            \begin{tabular}{cccc}
                \toprule
                \textbf{Distillation Loss} & \textbf{Guidance} & \textbf{MAE} & \textbf{RMSE} \\
                \midrule
                 $\mathcal{L}_{\text{MSE}}$ & GD & 1.81 & 2.40 \\
                 $\mathcal{L}_{\text{MSE}}$ \& $\mathcal{L}_{\text{freq}}$                                   & GD & 1.79 & 2.36 \\
                 $\mathcal{L}_{\text{MSE}}$ \& $\mathcal{L}_{\text{grad}}$                                   & GD & 1.71 & 2.25 \\
                 $\mathcal{L}_{\text{MSE}}$ \& $\mathcal{L}_{\text{grad}}$ \& $\mathcal{L}_{\text{freq}}$  & GD & 1.69 & 2.23 \\
                 \midrule
                 $\mathcal{L}_{\text{MSE}}$ & CG & 1.25 & 1.71 \\
                 $\mathcal{L}_{\text{MSE}}$ \& $\mathcal{L}_{\text{grad}}$ \& $\mathcal{L}_{\text{freq}}$  & CG & 1.16 & 1.59 \\
                \bottomrule
            \end{tabular}
        }
    \end{minipage}

    \caption{\small Performance evaluation on the MIROC6 (2001) dataset. The left panel evaluates the impact of DDIM NFE to identify the optimal sweet spot (NFE=25). Based on this optimal trajectory, the right table presents an ablation study of student models distilled down to 6 steps, comparing different distillation losses and guidance strategies (GD vs. CG).}
    \label{fig_tab_combined}
    \vspace{-0.5cm} 
\end{figure}

\textbf{Impact of Distillation \& Conjugate Gradient Guidance.} 
To balance downscaling accuracy and inference efficiency, we first establish the performance upper bound of the teacher model by evaluating standard DDIM sampling on the MIROC6 dataset. As shown in Fig.~\ref{fig_tab_combined} (left), the 99\% quantile errors exhibit a clear U-shaped trend with respect to the NFE. The errors reach their minimum at NFE = 25 (MAE: 1.1434, RMSE: 1.6177), representing an optimal trade-off between detail generation and denoising stability. Notably, increasing sampling steps beyond 25 (e.g., to 50 or 200) yields no further improvements and even slightly degrades the performance. Consequently, we identify the NFE = 25 trajectory as the optimal trade-off to serve as the baseline for subsequent distillation.

Using the 25-step model as our teacher, we distill the generation process down to 6 steps for ablation analysis (Fig.~\ref{fig_tab_combined}, right). Results show that standard distillation at such a low-step regime yields limited accuracy. By explicitly introducing $\mathcal{L}_{\text{grad}}$ and $\mathcal{L}_{\text{freq}}$ into the training phase, we force the student to accurately mimic the teacher's structural and spectral variations, significantly reducing both MAE and RMSE. Finally, replacing the standard GD solver with CG provides a more precise and stable optimization, further minimizing the overall error. Note that we opt for a direct one-step distillation rather than standard progressive halving (e.g., $25 \rightarrow 12 \rightarrow 6$); detailed justifications and empirical results demonstrating their comparable performance are provided in Appendix~\ref{Further Ablation Study About Distillation}.

Overall, our two-stage evaluation paradigm demonstrates that blindly increasing sampling steps is computationally inefficient. By anchoring the distillation process to the empirically verified sweet spot (NFE = 25) and applying the enhanced loss alongside CG, our method achieves competitive downscaling accuracy in only 6 inference steps, offering a practical solution for efficient high-resolution meteorological data generation.

\section{Conclusion}

We introduced GCD, a GCM-consistent diffusion framework for zero-shot downscaling that is trained once on ERA5 and applied to heterogeneous CMIP6 GCMs without model-specific retraining. GCD conditions the diffusion prior on static boundaries (land--sea mask and elevation), which yields more terrain-consistent samples in our qualitative comparison. A spatial-domain filter measurement operator restricts guidance to scales of $5^\circ$ and above, which reduces the high-frequency spectral mismatch between GCM inputs and the ERA5-trained prior. To lower inference cost, we combine single-stage distillation ($25 \rightarrow 6$ steps) with conjugate gradient guidance, which reduced the 6-step error relative to gradient-descent guidance in our ablations. Across five GCMs, the full 1000-step GCD attains lower 99th-percentile errors than all compared baselines. The 6-step version does so on three of the five GCMs while running over $130\times$ faster than vanilla DPS, but it degrades on IPSL. The generated fields also show vortex structures consistent with tropical cyclones, which we assess distributionally rather than event by event. The assumptions and limits of these results are discussed in Section~\ref{sec:limitations}.

\section{Acknowledgements}
This work was supported by the Key Program of the National Natural Science Foundation of China (NSFC) (Grant No.~72234002), the National Natural Science Foundation of China (NSFC) (Grant No.~72504061), and the AI for Science Program, Shanghai Municipal Commission of Economy and Informatization (No.~2025-GZL-RGZN-BTBX-02031).

\bibliographystyle{unsrtnat}
\bibliography{example_paper}

\appendix

\newpage

\section{Limitations}
\label{sec:limitations}

GCD has several limitations.

\textbf{Static boundary conditions.} The geographically-constrained prior uses only static boundaries (land--sea mask and elevation). Dynamic physical constraints (e.g., time-varying ocean states) could further improve physical consistency and are left for future work.

\textbf{Inference cost.} The 6-step scheme reduces inference time substantially, but diffusion sampling remains more expensive than deterministic regression: GCD (6 NFE) takes 4.38 s per sample, versus 0.04 s for ESRGAN, 0.25 s for SwinIR, and 2.33 s for CorrDiff. This is a challenge for large ensemble generation. Distillation also adds one-time training (24.5 GPU hours in total).

\textbf{Linear measurement operators.} The CG solver relies on the linearity of $\mathcal{A}$. Nonlinear operators would require other solvers or standard DPS-style gradient guidance.

\textbf{Scales below $5^\circ$.} Guidance constrains only scales at or above $5^\circ$. Finer structure comes from the ERA5-trained prior and is not constrained by the GCM.

\textbf{No event-level validation.} GCMs are free-running and not temporally aligned with ERA5, so we evaluate 99th-percentile statistics and spectra and show tropical cyclones as qualitative cases. Our results do not establish that individual generated events are correct. Evaluation on nudged or reanalysis-driven simulations is a natural next step.

\textbf{Degradation on IPSL.} GCD (6 NFE) is worse than bilinear interpolation on IPSL (MAE 1.49 versus 1.06). The IPSL spectrum deviates strongly from ERA5 (Fig.~\ref{ablation3.1}(b)). We hypothesize that the 6-step student is less tolerant of this shift. The $\lambda$ and NFE sweeps in Appendix~\ref{app:ipsl} are consistent with this hypothesis, but we did not directly diagnose the input error by wavenumber band.

\textbf{Scope of experiments.} We evaluate five CMIP6 GCMs in the historical experiment and five variables. The teacher step budget (NFE=25) was selected on MIROC6 only. We have not tested future scenarios, other variables, or other GCMs. Appendix~\ref{app:generalization} reports NFE and $\lambda$ sensitivity on all five GCMs.

\section{Supplementary Theory and Methods}

\subsection{Supplementary Theory in Sec.~\ref{subsec:prior_and_measurement}}
\label{SI: Supplementary Theory 3.1}

The forward process diffuses the data distribution into Gaussian noise, while the reverse process learns to recover the data under the guidance of $\mathbf{C}$. The forward marginal distribution $q(\mathbf{X}_t|\mathbf{X}_0)$ and the conditional reverse transition $p_\theta(\mathbf{X}_{t-1}|\mathbf{X}_t, \mathbf{C})$ are parameterized as

{
\small
\begin{equation}
\begin{aligned}
    q(\mathbf{X}_t|\mathbf{X}_0) &= \mathcal{N}(\mathbf{X}_t; \sqrt{\bar{\alpha}_t}\mathbf{X}_0, (1 - \bar{\alpha}_t)\mathbf{I}), \\
    p_\theta(\mathbf{X}_{t-1}|\mathbf{X}_t, \mathbf{C}) &= \mathcal{N}(\mathbf{X}_{t-1}; \boldsymbol{\mu}_\theta(\mathbf{X}_t, t, \mathbf{C}), \beta_t \mathbf{I}), 
\end{aligned}
\end{equation}
}

where $\beta_t$ represents the pre-defined noise variance schedule, $\alpha_t \triangleq 1 - \beta_t$ , $\bar{\alpha}_t \triangleq \prod_{s=1}^t \alpha_s $. $\boldsymbol{\mu}_\theta$ denotes the reverse process mean function approximator and $\boldsymbol{\mu}_\theta(\mathbf{X}_t, t, \mathbf{C}) =  \frac{1}{\sqrt{\alpha_t}} (\mathbf{X}_t - \frac{\beta_t}{\sqrt{1-\bar{\alpha}_t}} \boldsymbol{\epsilon}_\theta(\mathbf{X}_t, t,\mathbf{C}))$. The network $\boldsymbol{\epsilon}_\theta$ is trained to approximate the Gaussian noise $\boldsymbol{\epsilon}$ added during the forward process by minimizing the variational lower bound:

{
\small
\begin{equation}
    \mathcal{L} = \mathbb{E}_{\mathbf{X}_0, \boldsymbol{\epsilon}, t, \mathbf{C}} \left[ \| \boldsymbol{\epsilon} - \boldsymbol{\epsilon}_\theta(\mathbf{X}_t, t, \mathbf{C}) \|^2_F \right]. \label{eq:variational lower bound}
\end{equation}
}

Let $\mathbf{X}_0 \sim q(\mathbf{X}_0)$ denote the high-fidelity atmospheric state variables sampled from the reanalysis data distribution. We formulate the prior using a conditional Denoising Diffusion Probabilistic Model (DDPM).

\subsection{Inference Algorithm of GCD}
\label{Inference Algorithm of ZSSD}

From the noisy state $\mathbf{X}_t$ using Tweedie’s formula \cite{NEURIPS2021_077b83af,Efron01122011}, we have the posterior mean:

{
\small
\begin{equation}
\hat{X}_0(\mathbf{X}_t, \mathbf{C}) := \mathbb{E}[\mathbf{X}_0 | \mathbf{X}_t,\mathbf{C}] = \frac{\mathbf{X}_t - \sqrt{1-\bar{\alpha}_t}\boldsymbol{\epsilon}_\theta(\mathbf{X}_t, t, \mathbf{C})}{\sqrt{\bar{\alpha}_t}}.  \label{eq:posterior mean}
\end{equation}
}

However, using only Eq.~\eqref{eq:posterior mean} for climate downscaling will result in inconsistencies with low-resolution observations and other issues. Therefore, posterior sampling based on Bayes' rule is required. We have:

{
\small
\begin{equation}
\log p(\mathbf{X}_t | \mathbf{Y},\mathbf{C}) \propto \log p(\mathbf{X}_t|\mathbf{C}) + \log p(\mathbf{Y} | \mathbf{X}_t,\mathbf{C}). \label{eq:posterior sampling}
\end{equation}
}

Taking the gradient with respect to $\mathbf{X}_t$ to obtain the conditional score function:

{
\small
\begin{equation}
\nabla_{\mathbf{X}_t} \log p(\mathbf{X}_t | \mathbf{Y},\mathbf{C}) = \nabla_{\mathbf{X}_t} \log p(\mathbf{X}_t|\mathbf{C}) + \nabla_{\mathbf{X}_t} \log p(\mathbf{Y} | \mathbf{X}_t,\mathbf{C}).  \label{eq:conditional score function}
\end{equation}
}

In the framework of diffusion models, the noise-prediction network $\boldsymbol{\epsilon}_\theta(\mathbf{X}_t, t)$ is essentially equivalent to the time-dependent score function $\nabla_{\mathbf{X}_t} \log p(\mathbf{X}_t)$ \cite{6795935}. Therefore, once $\boldsymbol{\epsilon}_{\theta^*}$ is acquired by minimizing the variational lower bound in Eq.~\eqref{eq:variational lower bound}, we can use the approximation $\nabla_{\mathbf{X}_t} \log p(\mathbf{X}_t|\mathbf{C}) \simeq -\boldsymbol{\epsilon}_{\theta^*}(\mathbf{X}_t, t, \mathbf{C})/\sqrt{1 - \bar{\alpha}_t}$, and the approximation error arises from an optimization error with respect to Eq.~\eqref{eq:variational lower bound}. 

The second term $\nabla_{\mathbf{X}_t} \log p(\mathbf{Y} | \mathbf{X}_t,\mathbf{C})$ in Eq.~\eqref{eq:conditional score function} is intractable, but according to $p(\mathbf{Y}|\mathbf{X}_t,\mathbf{C}) = \mathbb{E}_{\mathbf{X}_0 \sim p(\mathbf{X}_0|\mathbf{X}_t,\mathbf{C})} [p(\mathbf{Y}|\mathbf{X}_0)]$ and $\hat{\mathbf{X}}_0 := \mathbb{E}[\mathbf{X}_0 | \mathbf{X}_t,\mathbf{C}] = \mathbb{E}_{\mathbf{X}_0 \sim p(\mathbf{X}_0 | \mathbf{X}_t,\mathbf{C})} [\mathbf{X}_0]$, it can be approximated by conditioning on the estimated clean image $\hat{\mathbf{X}}_0$:

{
\small
\begin{equation}
p(\mathbf{Y} | \mathbf{X}_t,\mathbf{C}) \simeq p(\mathbf{Y} | \hat{\mathbf{X}}_0(\mathbf{X}_t,\mathbf{C})),  \label{eq:approximated estimated clean image}
\end{equation}
}

implying that the outer expectation of $p(\mathbf{Y}|\mathbf{X}_0)$ over the posterior distribution is replaced with the inner expectation of $\mathbf{X}_0$. Thus, the approximation error in Eq.~\eqref{eq:approximated estimated clean image} can be quantified with the Jensen gap \cite{gao2020boundsjensengapimplications,SIMIC2008414}. Furthermore, this Jensen’s gap has an exact upper bound under our inverse problem subject to Gaussian noise~\cite{chung2022diffusion}.

To explicitly formulate this likelihood for our downscaling task, we consider the common measurement model stated in Eq.~\eqref{eq:degradation_model}. In our framework, $\boldsymbol{\Sigma}_W$ is set to $\mathbf{I}$ by assuming longitudinal homoscedasticity, as the longitudinal span of grid cells remains invariant across the domain. Therefore, the log-likelihood is instantiated as follows:

{
\small
\begin{equation}
\log p(\mathbf{Y} | \hat{\mathbf{X}}_0(\mathbf{X}_t,\mathbf{C})) \simeq -  \| \mathbf{Y} - \mathcal{A}_\text{filter}(\hat{\mathbf{X}}_0(\mathbf{X}_t,\mathbf{C})) \|^2_{\mathbf{\Lambda},\mathbf{I}}, \label{eq:log-likelihood}
\end{equation}
}

where $\mathbf{\Lambda} \triangleq \mathbf{\Sigma}^{-1}_H$,  $\|\mathbf{X}\|^2_{\mathbf{\Lambda},\mathbf{I}} \triangleq \text{Tr} (\mathbf{\Lambda}\mathbf{X}\mathbf{X}^{\top})$. The formulation of the above is derived from the density function for the  Matrix Variate Gaussian (MVG) distribution \cite{gupta1999matrix}.

Then, we rewrite the log-likelihood in Eq.~\eqref{eq:log-likelihood} to quantify the distance between $\mathcal{A}_\textbf{filter}(\hat{\mathbf{X}}_0)$ and $\mathbf{Y}$. We also incorporate latitude weighting into this distance calculation:

{
\small
\begin{equation}
    \mathcal{E}(\mathbf{X}_t, \mathbf{Y}) = \| \mathbf{W}  (\mathcal{A}_\textbf{filter}(\hat{\mathbf{X}}_0(\mathbf{X}_t, \mathbf{C})) - \mathbf{Y}) \|^2_F, \label{eq:latitude weighting into distance}
\end{equation}
}

where $\mathbf{W}^{\top}\mathbf{W}=\mathbf{\Sigma}^{-1}_H$. 

Notably, the distance in Eq.~\eqref{eq:latitude weighting into distance} differs from the log-likelihood in Eq.~\eqref{eq:log-likelihood} by only one negative sign. The weighting matrix $\mathbf{W}$ combines the spherical geometry of the Earth and the measurement noise at different latitudes, with weights expressed as $\mathbf{W}_{ii} =\sqrt{\mathbf{\Sigma}^{-1}_{H_{ii}}} \propto \cos(\phi_i)$, where $\phi_i$ is the latitude of the $i$-th grid row.

Finally, we use the gradient $\nabla_{\mathbf{X}_t} \mathcal{E}$ to guide the generation process. This ensures that the generated high-resolution data is consistent with the low-resolution input. Specifically, when downsampled back to the coarse scale, the generated output matches the original GCMs projection while preserving fine-scale variability. This approach preserves the fidelity of large-scale structures while augmenting realistic local details. The algorithm of inference refer to Algorithm~\ref{alg:sampling}.

\subsection{Accelerated Inference of GCD}
\label{Inference Algorithm of Conjugate Gradient Acceleration}

\begin{figure}[t]
  \centering
  \begin{minipage}[t]{0.48\textwidth}
    \begin{algorithm}[H]
      \caption{GCD inference}
      \label{alg:sampling}
      \begin{algorithmic}[1]
        \STATE {\bfseries Require:} $T, \mathbf{Y}_{\text{raw}}, \mathbf{C}, \zeta, \mathbf{W}$
        \STATE {\bfseries Init:} $\mathbf{X}_T \sim \mathcal{N}(\mathbf{0}, \mathbf{I})$, $\mathbf{Y} \leftarrow \mathcal{A}_{\text{filter}}(\mathbf{Y}_{\text{raw}})$
        \FOR{$t=T-1$ {\bfseries to} $0$}
          \STATE $\mathbf{Z} \sim \mathcal{N}(\mathbf{0}, \mathbf{I})$
          \STATE $\boldsymbol{\epsilon} \leftarrow \boldsymbol{\epsilon}_\theta(\mathbf{X}_t, t, \mathbf{C})$
          \STATE $\hat{\mathbf{X}}_0 \leftarrow \frac{1}{\sqrt{\bar{\alpha}_t}}(\mathbf{X}_t - \sqrt{1-\bar{\alpha}_t}\boldsymbol{\epsilon})$
          \STATE $\mathbf{G} \leftarrow \nabla_{\mathbf{X}_t} \| \mathbf{W} (\mathcal{A}_{\text{filter}}(\hat{\mathbf{X}}_0) - \mathbf{Y}) \|^2_F$
          \STATE $\boldsymbol{\mu}_t \leftarrow \frac{1}{\sqrt{\alpha_t}} \big( \mathbf{X}_t - \frac{\beta_t}{\sqrt{1-\bar{\alpha}_t}} \boldsymbol{\epsilon} \big)$
          \STATE $\mathbf{X}_{t-1} \leftarrow \boldsymbol{\mu}_t + \sqrt{\beta_t} \mathbf{Z} - \zeta \mathbf{G}$
        \ENDFOR
        \STATE \textbf{return} $\hat{\mathbf{X}}_0$
      \end{algorithmic}
    \end{algorithm}
  \end{minipage}
  \hfill
  \begin{minipage}[t]{0.48\textwidth}
    \begin{algorithm}[H]
      \caption{GCD inference acceleration}
      \label{alg:distilled_dps}
      \begin{algorithmic}[1]
        \STATE {\bfseries Require:} $K, \mathbf{Y}_{\text{raw}}, \mathbf{C}, \{t_k\}_{k=1}^K, \tau, \lambda, M$
        \STATE {\bfseries Init:} $\mathbf{X}_{t_{K}} \sim \mathcal{N}(\mathbf{0}, \mathbf{I})$, $\mathbf{Y} \leftarrow \mathcal{A}_{\text{filter}}(\mathbf{Y}_{\text{raw}})$
        \FOR{$k=K-1$ {\bfseries to} $0$}
          \STATE $\boldsymbol{\epsilon} \leftarrow \boldsymbol{\epsilon}_\psi(\mathbf{X}_{t_k}, t_k, \mathbf{C})$
          \STATE $\hat{\mathbf{X}}_0 \leftarrow \frac{1}{\sqrt{\bar{\alpha}_{t_k}}} \big(\mathbf{X}_{t_k} - \sqrt{1-\bar{\alpha}_{t_k}}\boldsymbol{\epsilon}\big)$
          \STATE $\mathbf{X}_{t_{k-1}} \leftarrow \sqrt{\bar{\alpha}_{t_{k-1}}}\hat{\mathbf{X}}_0 + \sqrt{1-\bar{\alpha}_{t_{k-1}}}\boldsymbol{\epsilon}$
          \STATE $\mathbf{X}_0^{(0)} \leftarrow \hat{\mathbf{X}}_0$ \textit{\# Initialize CG}
          \FOR{$m=1$ {\bfseries to} $M$}
            \STATE $\mathbf{X}_0^{(m)} \leftarrow \text{CG\_Step}(\mathbf{X}_0^{(m-1)}, \mathbf{Y}, \lambda, \tau)$
          \ENDFOR
          \STATE $\mathbf{X}_0^\star \leftarrow \mathbf{X}_0^{(M)}$
          \STATE $\mathbf{X}_{t_{k-1}} \leftarrow \mathbf{X}_{t_{k-1}} + \sqrt{\bar{\alpha}_{t_{k-1}}}(\mathbf{X}_0^\star-\hat{\mathbf{X}}_0)$
        \ENDFOR
        \STATE \textbf{return} $\mathbf{X}_{t_0}$
      \end{algorithmic}
    \end{algorithm}
  \end{minipage}
\end{figure}

Algorithm \ref{alg:distilled_dps} outlines the accelerated inference scheme of the GCD framework, which effectively compresses the generation trajectory by integrating progressive distillation with Conjugate Gradient (CG) guidance. The algorithm requires several specific inputs for the few-step regime: the reduced total number of sampling steps $K$ (e.g., $K=6$), the raw coarse observation $\mathbf{Y}_{\text{raw}}$, the physical conditioning context $\mathbf{C}$, and the discrete time schedule $\{t_k\}_{k=1}^K$ tailored for the distilled model. Furthermore, it takes hyperparameters for the inner CG solver, including the number of inner iterations $M$, the regularization weight $\lambda$, and the solver tolerance $\tau$.

Similar to the standard procedure, the process initializes with a pure noise state $\mathbf{X}_{t_{K}} \sim \mathcal{N}(\mathbf{0}, \mathbf{I})$ and extracts the smoothed observation $\mathbf{Y} = \mathcal{A}_{\text{filter}}(\mathbf{Y}_{\text{raw}})$. The reverse generative process iterates over the compressed schedule from $k=K-1$ down to $0$. At each step, the distilled student model $\boldsymbol{\epsilon}_\psi$ rapidly predicts the noise component, enabling the estimation of the clean high-resolution state $\hat{\mathbf{X}}_0$ across a large time stride. Based on this uncorrected estimate, a preliminary next state $\mathbf{X}_{t_{k-1}}$ is computed. 

However, because aggressive large-step sampling inevitably introduces significant truncation errors and deviates from the data manifold, an inner optimization loop is introduced to improve data consistency. Starting with the initial guess $\mathbf{X}_0^{(0)} = \hat{\mathbf{X}}_0$, the algorithm executes $M$ iterations of the `CG\_Step`. This inner solver efficiently traverses the current data manifold to solve the optimization problem associated with the inverse problem (Eq. \eqref{eq: optimization objective of CG}). This iterative refinement yields an optimized state $\mathbf{X}_0^\star$ that is consistent with the coarse input. Finally, to correct the diffusion trajectory, the scaled difference between the optimized state and the original estimate, $\sqrt{\bar{\alpha}_{t_{k-1}}}(\mathbf{X}_0^\star-\hat{\mathbf{X}}_0)$, is injected back into the preliminary state $\mathbf{X}_{t_{k-1}}$. This dual-stage mechanism ensures that the ultra-fast prior transitions provided by distillation are precisely calibrated by the CG guidance, resulting in high-fidelity downscaling in minimal steps.

\subsection{Supplementary Evaluation Methods}
\label{Supplementary Evaluation Methods}

Climate downscaling differs from short-term weather prediction in evaluation method.
Although GCMs and ERA5 reanalysis share nominal timestamps in the historical period, their internal weather realizations are not phase-aligned, rendering pointwise time-synchronous comparison statistically inappropriate. Consequently, evaluation focuses on distributional consistency and extreme-state characteristics under comparable climatological conditions.
To evaluate the model’s ability to reproduce extreme climate states, we analyze the spatial statistics of upper-tail behavior.

We define the MAE and RMSE at the $99^{th}$ percentile as follows:
{\small
\begin{equation}
\text{MAE} = \frac{1}{W \sum_{j=1}^{H} w_j} \sum_{i=1}^{W} \sum_{j=1}^{H} w_j \left| Q_{\text{model},i,j} - Q_{\text{ERA5},i,j} \right|,
\label{eq:mae_99}
\end{equation}
}

{\small
\begin{equation}
\text{RMSE} = \sqrt{ \frac{1}{W \sum_{j=1}^{H} w_j} \sum_{i=1}^{W} \sum_{j=1}^{H} w_j \left( Q_{\text{model},i,j} - Q_{\text{ERA5},i,j} \right)^2 },
\label{eq:rmse_99}
\end{equation}
}

where $W=1440$ and $H=720$ denote the number of grid points in the longitude and latitude directions, respectively. $Q_{\text{model},i,j}$ and $Q_{\text{ERA5},i,j}$ represent the 99th percentile values at grid point $(i,j)$. Crucially, $w_j = \cos(\phi_j)$ represents the latitude weight, which corrects for the area distortion inherent in the spherical geometry of the Earth.

\section{Supplementary Experiments Details}

\subsection{Details of Datasets}
\label{Details of Datasets}

\begin{wraptable}{r}{0.55\textwidth} 
   \vspace{-10pt}
  \caption{Summary of GCMs used in this study.}
  \vspace{-8pt}
  \label{tab:cmip6_models}
  \begin{center}
    \begin{footnotesize} 
      \begin{sc}
        \setlength{\tabcolsep}{3.5pt}
        \begin{tabular}{llcr}
          \toprule
          Model Name      & Institution & Nom. Res. & Grid Size \\
          \midrule
          IPSL-CM6A-LR    & IPSL        & $\sim$250 km & 144 $\times$ 143 \\
          AWI-ESM-1-1-LR  & AWI         & $\sim$250 km & 192 $\times$ 96  \\
          MPI-ESM1-2-LR   & MPI-M       & $\sim$250 km & 192 $\times$ 96  \\
          MIROC6          & JAMSTEC     & $\sim$140 km & 256 $\times$ 128 \\
          MPI-ESM1-2-HR   & MPI-M       & $\sim$100 km & 384 $\times$ 192 \\
          \bottomrule
        \end{tabular}
      \end{sc}
    \end{footnotesize} 
  \end{center}
  \vspace{-20pt}
\end{wraptable}

As shown in Table~\ref{tab:cmip6_models}, these models cover a range of resolutions from about 100 km to 250 km. 
This diversity enables an assessment of whether our method generalizes across different grid resolutions without model-specific retraining.

To provide a clear and challenging evaluation of downscaling performance, we focus on TCs–related atmospheric fields as a representative case. Specifically, we select five variables: Mean Sea Level Pressure (MSL) and 10~m Wind Components (U10M, V10M) to characterize near-surface intensity, and Geopotential Height at 500hPa and 250hPa (Z500, Z250) to represent upper-level circulation patterns.

We select the same variables as in the ERA5 dataset. In CMIP6, these are named `psl' (MSL), `uas' and `vas' (10m wind), and `zg' (geopotential height). Specifically, we multiply the `zg' fields at 500hPa and 250hPa by the gravitational acceleration $g$ to convert them into geopotential (Z), ensuring consistency with the ERA5 format.

\subsection{Settings Details of Baselines}
\label{Settings Details of Baselines}

In our evaluation, the standard supervised baselines (SwinIR, ESRGAN, CorrDiff, CM) are trained on artificially degraded ERA5 data. Since their inputs must strictly align with the spatial characteristics of the training distribution, applying them directly to unseen heterogeneous GCMs constitutes a direct-transfer task. We use this setting to explicitly demonstrate the sub-optimal cross-domain performance caused by domain shifts when model-specific retraining is omitted. For CM+noise, we follow \citet{hess2025fast} by injecting statistical noise during training to mitigate these domain gaps.

\textbf{SwinIR}. The network was configured with a window size of $8 \times 8$, 4 Residual Swin Transformer Blocks (RSTB), 4 Swin Transformer Layers (STL) per block, and 128 channels. Training was conducted using the AdamW optimizer ($\beta_1=0.9, \beta_2=0.999$) with an initial learning rate of $2 \times 10^{-4}$ and an $\mathcal{L}_1$ pixel loss.

\textbf{ESRGAN}. The generator utilizes 12 Residual-in-Residual Dense Blocks (RRDB) with 64 channels. The model was optimized using Adam with a learning rate of $1 \times 10^{-4}$. The total loss function was formulated as a weighted sum of the $\mathcal{L}_1$ pixel loss ($\lambda=1 \times 10^{-2}$), a VGG-based perceptual loss ($\lambda=1.0$), and a relativistic adversarial loss ($\lambda=5 \times 10^{-3}$).

\textbf{CorrDiff}. The architecture comprises two distinct networks: a deterministic regression UNet and a generative diffusion UNet. Both networks share an identical backbone configuration with channel dimensions $(64, 128, 256, 512)$, though they are optimized independently. The regression UNet is first trained using an $\mathcal{L}_2$ loss to predict the smoothed high-resolution mean directly from the coarse input. Subsequently, the diffusion UNet is trained to generate the high-frequency residual, conditioned on the coarse input. The diffusion model was trained with a linear noise schedule over $T=1000$ steps. During inference, we utilized the DDIM sampler for 12 NFEs to balance computational efficiency and sample quality.

\textbf{CM and CM+NOISE}. The baseline Consistency Model (CM) was trained using the continuous-time formulation with a maximum time boundary of $T=80$ and an EMA decay rate of $0.999$. The optimization objective combined an $\mathcal{L}_2$ pixel loss with an LPIPS perceptual metric. To maximize inference efficiency, sampling was executed using the single-step consistency sampler (1 NFE). For the CM+NOISE variant, we followed the exact same architecture but introduced a data augmentation strategy to simulate domain shifts. Specifically, during training, artificial noise was injected into the coarse-resolution ERA5 inputs by applying the standard DDPM forward noise scheduler for $t=50$ steps. Correspondingly, during the inference phase, this identical 50-step DDPM noise degradation was applied to the raw GCM inputs prior to single-step generation.

\textbf{DDRM}. Following the original implementation, all experiments were performed with the default settings of $\eta_B = 1.0$ and $\eta = 0.85$. We leveraged the DDIM sampler for 20 NFEs. The forward measurement operator was implemented using 2D average pooling to construct the singular value decomposition (SVD) matrices efficiently.

\textbf{Vanilla DPS}. We utilized the same backbone without the geographic and temporal conditioning. Inference was performed using the standard DDPM sampler for the full 1000 NFEs. The gradient guidance scale was empirically set to $\zeta = 0.5$, and the measurement operator was defined as standard spatial average pooling without any frequency-domain filtering.

\subsection{Settings Details of GCD}
\label{Settings Details of ZSSD}

The backbone of our pre-trained prior is a 2D U-Net architecture with a four-level encoder--decoder structure and skip connections to preserve multi-scale spatial information. The channel dimensions for the downsampling and upsampling paths are set to $\{64, 128, 256, 512\}$, respectively. Crucially, the bottleneck stage incorporates cross-attention mechanisms to explicitly integrate temporal embeddings, conditioning the spatial features on time information. Each resolution level comprises two stacked residual blocks to enhance feature extraction capabilities. 

The model was pre-trained and inferred on a cluster of 8 NVIDIA H200 GPUs (141~GB memory per GPU). We used the AdamW optimizer with an initial learning rate of $10^{-4}$. A StepLR scheduler with decays factor  $\gamma=0.9$ was applied every 50 epochs. Training was conducted for 75 epochs. Subsequently, the distillation model was initialized from this pre-trained model and trained for an additional 30 epochs, maintaining the exact same learning rate and scheduler settings. We empirically set the balancing weights to $ \lambda_{\text{grad}} = 0.1 $ and $ \lambda_{\text{freq}} = 0.01 $ in distill loss function. 

\section{Supplementary Results}
\subsection{Supplementary Results in Main Results}
\label{Supplementary Results in main results}
\begin{figure}[!htbp]
  \begin{center}
    \centerline{\includegraphics[width=\textwidth]{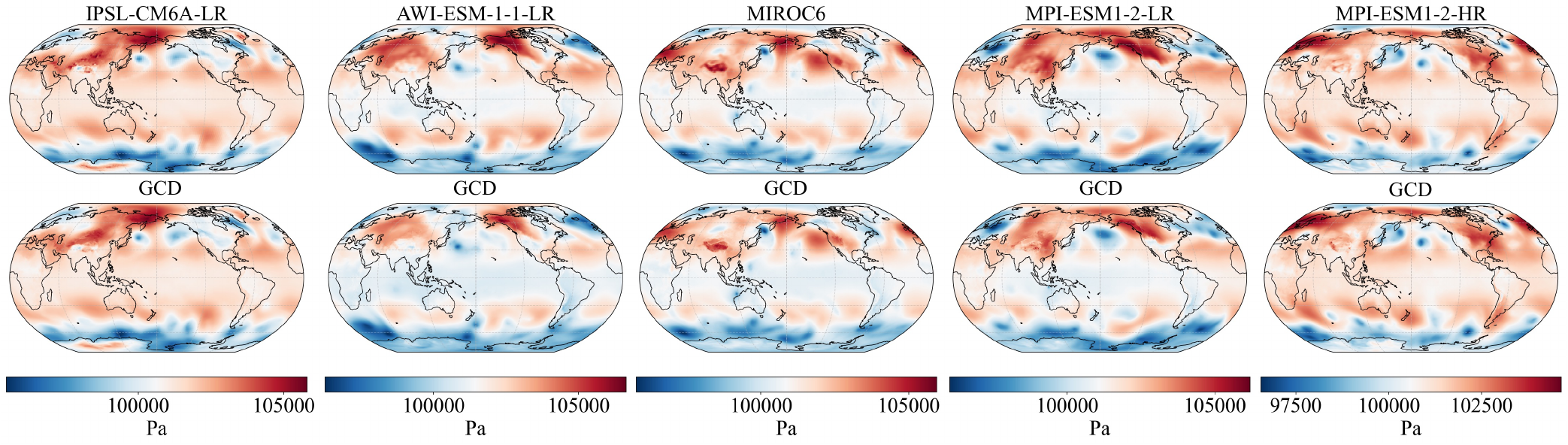}}
    \caption{
Visualization of MSL fields across different GCMs before and after downscaling. The time period is consistent with Fig.~\ref{fig:tc_vis}. Note that unlike surface pressure, which mirrors rugged topography, MSL is a derived variable designed to filter out orographic effects, resulting in a naturally smoother field at synoptic scales, although major features like the Tibetan Plateau and the Andes remain visually discernible. To quantify the variations that are less apparent, we provide a spectral analysis in Fig.~\ref{appendix_psd} to illustrate the differences in small-scale energy.
    }
    \label{appendix_msl}
  \end{center}
  \vspace{-0.5cm}
\end{figure}

\begin{figure}[!htbp]
  \begin{center}
    \centerline{\includegraphics[width=\textwidth]{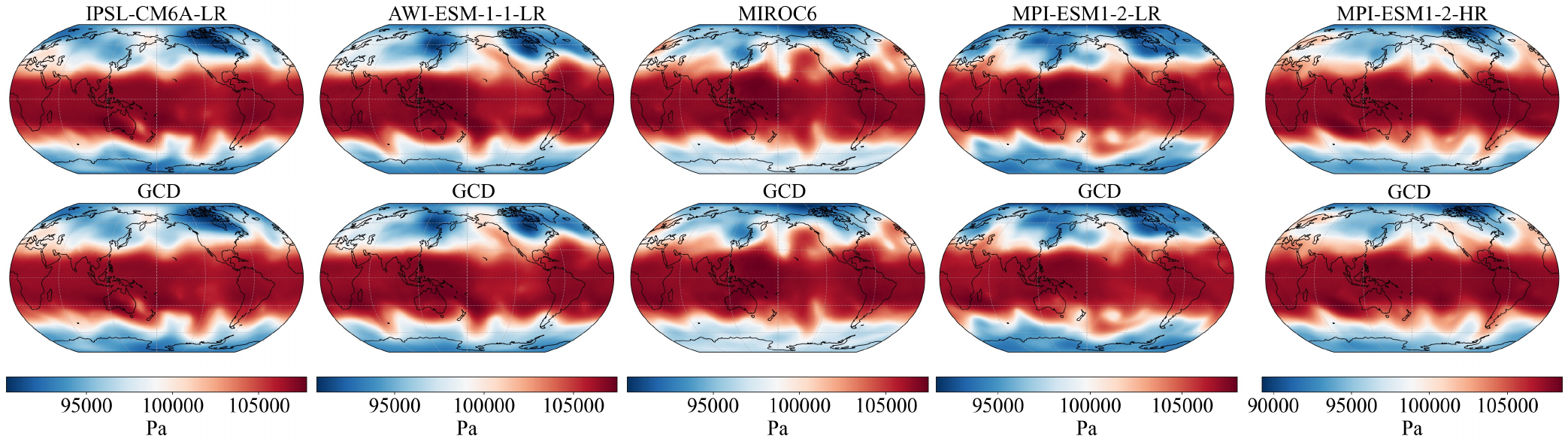}}
    \caption{
Visualization of Z250 fields across different GCMs before and after downscaling. The time period is consistent with Fig.~\ref{fig:tc_vis}. Since upper-level fields are generally smooth, making differences in small-scale details difficult to distinguish visually, we conducted a spectral analysis in Fig.~\ref{appendix_psd}.
    }
    \label{appendix_z250}
  \end{center}
  \vspace{-1.5cm}
\end{figure}

\begin{figure}[!htbp]
  \vspace{-0.5cm}
  \begin{center}
    \centerline{\includegraphics[width=\textwidth]{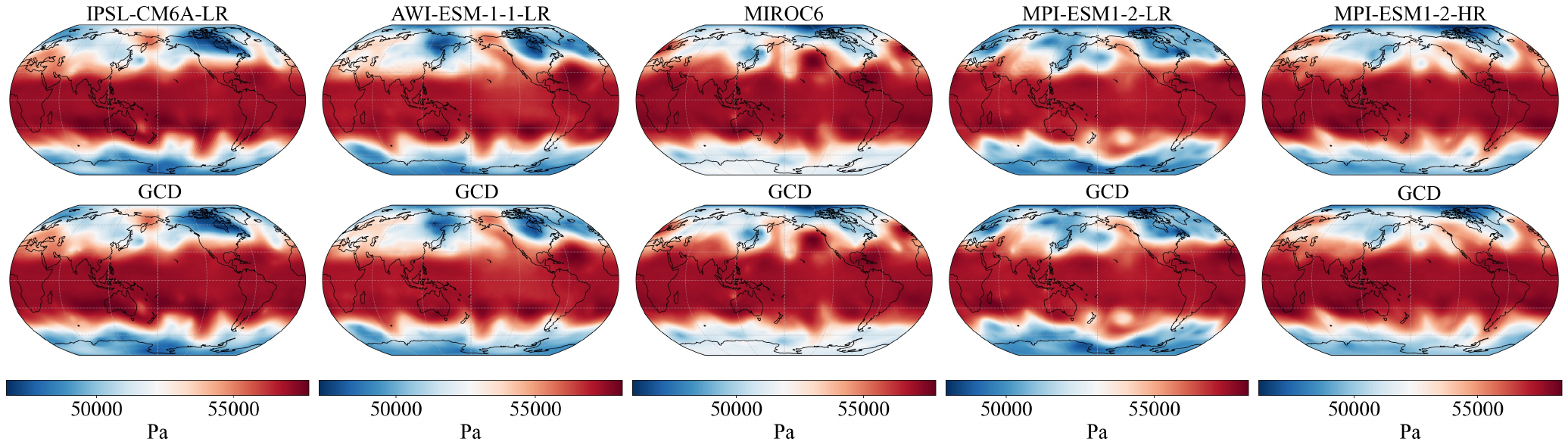}}
    \caption{
Visualization of Z500 fields across different GCMs before and after downscaling. The time period is consistent with Fig.~\ref{fig:tc_vis}. Since upper-level fields are generally smooth, making differences in small-scale details difficult to distinguish visually, we conducted a spectral analysis in Fig.~\ref{appendix_psd}.
    }
    \label{appendix_z500}
  \end{center}
  \vspace{-0.5cm}
\end{figure}

\begin{figure}[!htbp]
  \begin{center}
    \centerline{\includegraphics[width=\textwidth]{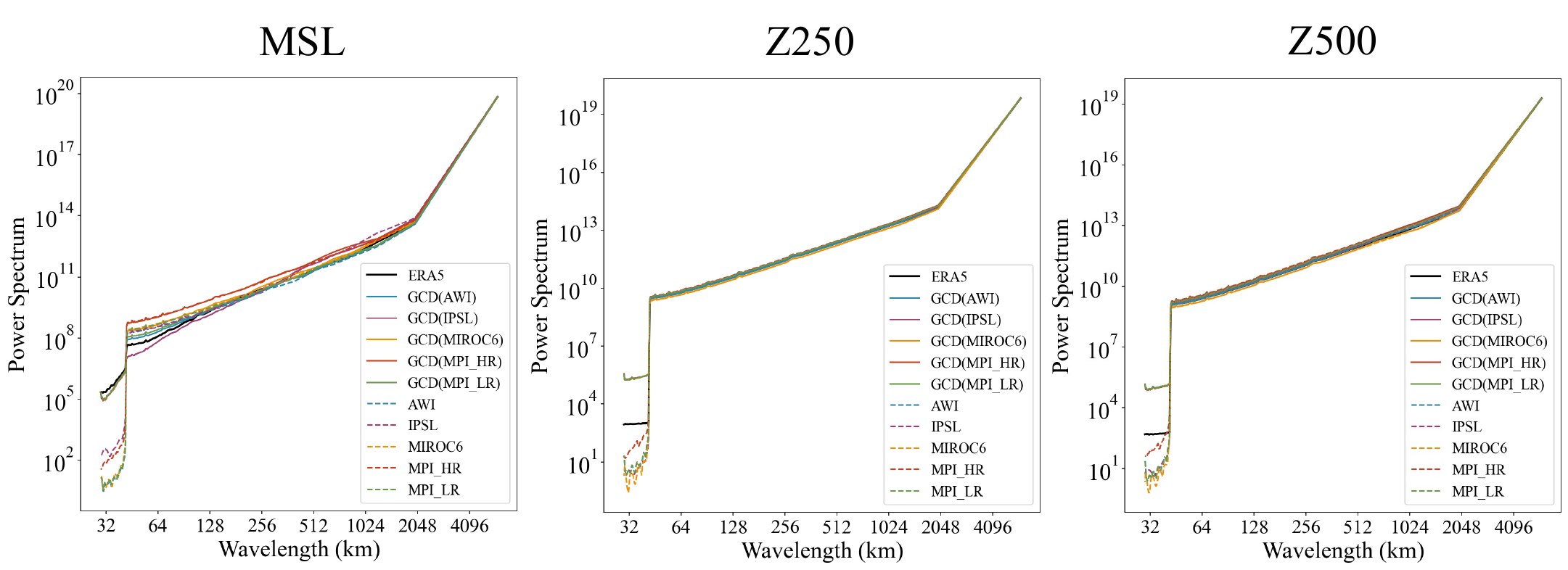}}
    \caption{
PSD of MSL, Z250, Z500 fields across different GCMs before and after downscaling. The time period is consistent with Fig.~\ref{fig:tc_vis}.
    }
    \label{appendix_psd}
  \end{center}
  \vspace{-0.3cm}
\end{figure}

This section complements Fig.~\ref{fig:tc_vis} by further illustrating the downscaling results for pressure-related variables. Since upper-level fields are generally smooth, the visual differences before and after downscaling are not significant for Z250 and Z500 (see Fig.~\ref{appendix_z250} and Fig.~\ref{appendix_z500}). In contrast, MSL shows clear improvements in detail over regions like the Tibetan Plateau and the Andes (see Fig.~\ref{appendix_msl}). To capture nuances beyond visual inspection, we further analyzed the spectral characteristics (Fig.~\ref{appendix_psd}). In the frequency domain, distinct behaviors emerge at small scales: for MSL, GCMs tend to overestimate high-frequency energy, whereas our results align consistently with ERA5. Conversely, for Z250 and Z500, where the small-scale energy in GCMs nearly vanishes, our method effectively recovers these high-frequency components, exhibiting slightly higher energy levels than ERA5. Despite the limited visual change, it is important to note that Z250 and Z500 remain crucial for the accurate localization and identification of TCs.

\begin{figure}[!htbp]
    \centering
    \includegraphics[width=\textwidth]{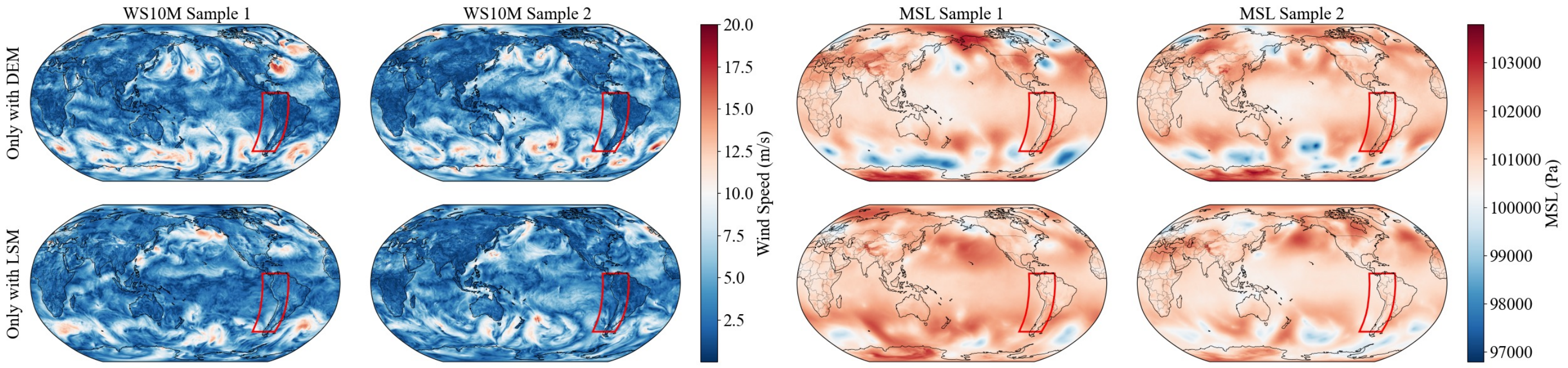}
    \caption{Ablation study of individual geographical constraints. Both the ``Only with DEM'' and ``Only with LSM'' configurations successfully capture the terrain blocking effect over the Andes (red boxes in WS10M) and maintain stable MSL patterns.}
    \label{fig:ablation_dem_lsm}
    \vspace{-0.3cm}
\end{figure}

\subsection{Further Ablation Study About Prior}
\label{Further Ablation Study About Prior}

In the main text, we demonstrated that geographical constraints are essential for correcting unphysical artifacts. To isolate their individual contributions, we further evaluate the model conditioned on only the Digital Elevation Model (DEM) or only the Land Sea Mask (LSM). 

As shown in Figure~\ref{fig:ablation_dem_lsm}, both individual constraints successfully reproduce the terrain blocking effect over the Andes (red boxes in WS10M). Naturally, the DEM constraint works by providing explicit topographic heights to guide wind deceleration. Interestingly, the LSM constraint alone achieves a similar blocking effect. Because the Andes mountain range tightly follows the South American coastline, this sharp land-sea boundary acts as an effective spatial proxy for the mountain barrier. 

Similarly, for the Mean Sea Level pressure (MSL), both individual constraints independently suppress the spurious high-pressure anomalies observed in the unconstrained baseline. These results confirm that while DEM and LSM are individually robust geographical priors, combining them (as in our final model) ensures the most comprehensive physical fidelity.

\begin{wrapfigure}{r}{0.5\textwidth}
  \vspace{-0.15in} 
  \centering
  \includegraphics[width=\linewidth]{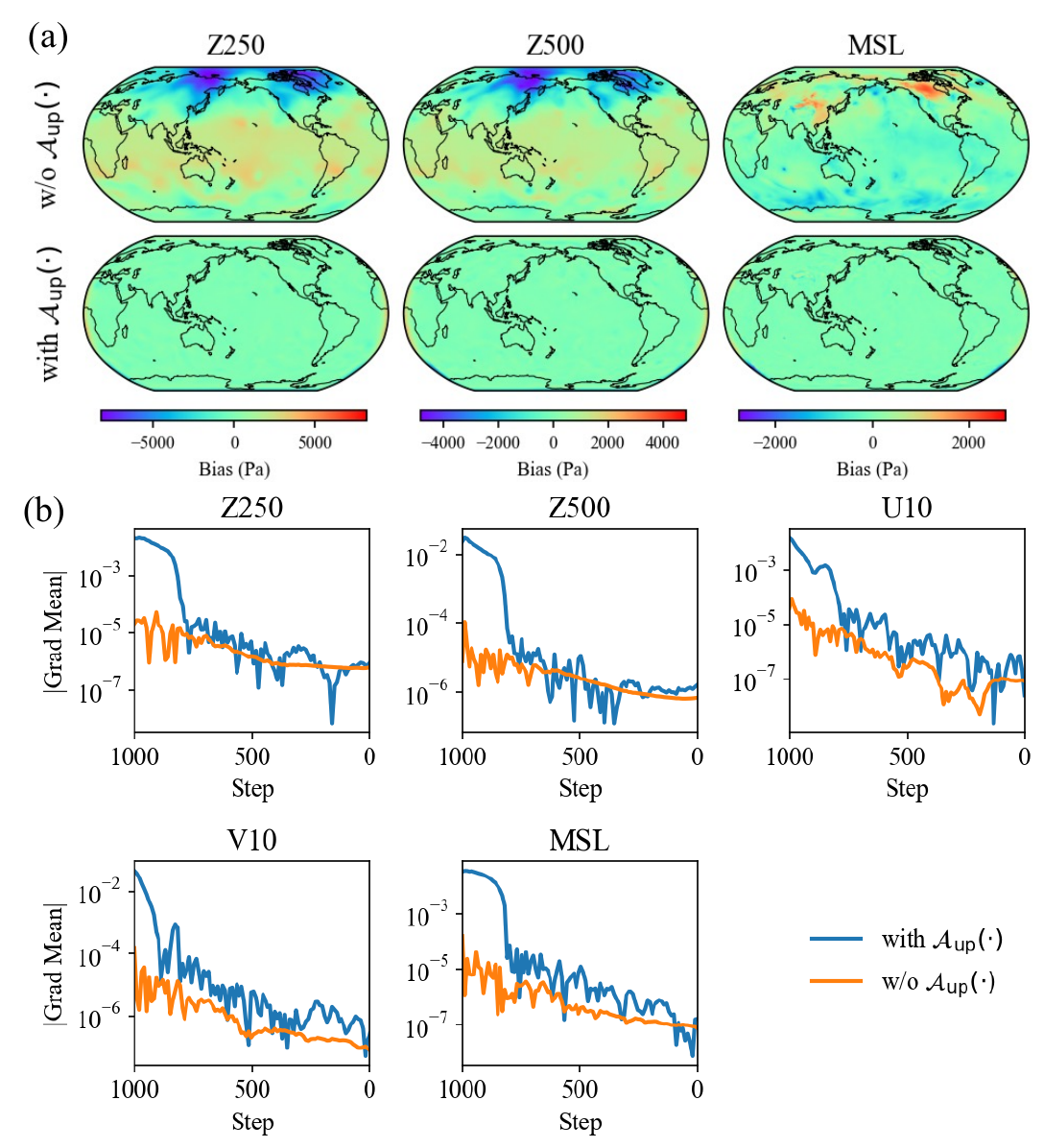}
  \caption{
    Impact of $\mathcal{A}_{\text{up}}(\cdot)$ in $\mathcal{A}_{\text{filter}}(\cdot)$. (a) Bias distributions. (b) Evolution of gradients during diffusion posterior sampling, illustrating the trajectory from timestep $t=1000$ to $0$ for both methods.
  }
  \label{ablation2.1}
  \vspace{-0.2in} 
\end{wrapfigure}

\subsection{Further Ablation Study About Measurement Operator}
\label{Further Ablation Study About Measurement Operator}

To investigate the advantages of $\mathcal{A}_{\text{filter}}$ over the vanilla downsampling operator ($\mathcal{A}_{\text{down}}$) in conditional guidance, we conducted an ablation study using 5° ERA5 data as the guidance signal. Under this setting, the core difference between the two operators lies in the presence or absence of the upsampling operator, $\mathcal{A}_{\text{up}}(\cdot)$. 

As shown in Fig.~\ref{ablation2.1}(a), removing $\mathcal{A}_{\text{up}}(\cdot)$ causes the model to lose effective constraints on large-scale physical fields, resulting in massive spatial biases in key variables such as Z250 and Z500. Conversely, incorporating $\mathcal{A}_{\text{up}}(\cdot)$ significantly stabilizes the spatial performance, nearly eliminating the biases.

To uncover the underlying mechanism, we tracked the evolution of gradients during the diffusion posterior sampling (DPS) process. As illustrated in Fig.~\ref{ablation2.1}(b), during the early stages of denoising (i.e., timestep $t \approx 1000$, when the model constructs macroscopic structures), the operator with $\mathcal{A}_{\text{up}}(\cdot)$ provides a sufficiently large and stable gradient guidance signal. In contrast, the absence of $\mathcal{A}_{\text{up}}(\cdot)$ leads to extremely small and noisy gradients early on, depriving the model of effective conditional constraints during the critical structure-formation phase, which ultimately causes the severe spatial distortion seen in (a). This clearly demonstrates the crucial role of $\mathcal{A}_{\text{filter}}$ in maintaining gradient stability and ensuring physical consistency.

\subsection{Further Ablation Study About Distillation}
\label{Further Ablation Study About Distillation}

Standard diffusion distillation often halves the step count iteratively (e.g., $25 \rightarrow 12 \rightarrow 6$). We instead distill directly ($25 \rightarrow 6$). To check whether this simplification hurts accuracy, we compare the two trajectories on MIROC6 (test year 2001). To isolate the effect of the trajectory, both students are trained with the same objective, $\mathcal{L}_{\text{MSE}}$ only, and evaluated with CG guidance at $NFE=6$. The absolute errors are therefore comparable to the $\mathcal{L}_{\text{MSE}}$ + CG row of Fig.~\ref{fig_tab_combined}, not to the full-loss row.

\begin{table}[h]
\centering
\caption{99th-percentile errors on MIROC6 (2001) for two distillation trajectories, both trained with $\mathcal{L}_{\text{MSE}}$ only and evaluated with CG guidance.}
\label{tab:distillation_ablation}
\begin{tabular}{lccc}
\toprule
Distillation Trajectory & Intermediate Models & MAE & RMSE \\
\midrule
$25 \rightarrow 12 \rightarrow 6$ (iterative halving) & 1 & 1.2523 & 1.7133 \\
$25 \rightarrow 6$ (direct, ours) & 0 & 1.2541 & 1.7135 \\
\bottomrule
\end{tabular}
\end{table}

The two trajectories give nearly identical errors ($\Delta\text{MAE} = 0.0018$, $\Delta\text{RMSE} = 0.0002$), and the direct trajectory needs no intermediate model. A plausible explanation is that iterative halving mainly helps when compressing hundreds of steps, whereas our teacher already uses only 25; we have not tested this explanation directly. We did not run the halving trajectory with the full loss $\mathcal{L}_{\text{MSE}}+\mathcal{L}_{\text{grad}}+\mathcal{L}_{\text{freq}}$, so the comparison does not cover that setting. We use the direct trajectory because it is simpler and cheaper to train at no measurable accuracy cost in this experiment.

\subsection{Full-distribution metrics}
\label{app:fulldist}

While the 99th-percentile MAE effectively captures the model's capability to reconstruct extreme weather events, evaluating the full distribution is essential to ensure that the generative process does not introduce systemic biases into normal climate states. Table~\ref{tab:app_fulldist} presents the Mean Absolute Error (MAE) for the mean field, as well as the 50th and 90th percentiles.

The results reveal the limitations of traditional interpolation methods in modeling tail distributions. For instance, on the MIROC6 dataset, while the Bilinear method achieves a reasonable mean field MAE of $0.76$, its absolute error widens significantly to $1.52$ at the 90th percentile. In contrast, GCD (1000 NFE) maintains high absolute accuracy across the distribution, with its 90th-percentile error ($0.88$) remaining highly competitive. Among all baselines, GCD (1000 NFE) has the lowest mean-field, P50 and P90 errors on all five GCMs.

\begin{table}[t]
\centering\small
\caption{MAE of the mean field, 50th and 90th percentile. Computed as for the 99th-percentile results in Table~\ref{tab:main_results}.}
\label{tab:app_fulldist}
\begin{tabular}{lccccc}
\toprule
Method & MIROC6 & IPSL & AWI & MPI-LR & MPI-HR \\
\midrule
\multicolumn{6}{l}{\emph{Mean field}}\\
Bilinear       & 0.76 & 0.48 & 0.68 & 0.65 & 0.59 \\
SwinIR         & 0.69 & 0.42 & 0.63 & 0.58 & 0.54 \\
ESRGAN         & 0.67 & 0.41 & 0.61 & 0.56 & 0.52 \\
CorrDiff       & 0.62 & 0.44 & 0.57 & 0.53 & 0.49 \\
CM+NOISE       & 0.59 & 0.39 & 0.55 & 0.50 & 0.46 \\
GCD (1000 NFE) & 0.49 & 0.36 & 0.51 & 0.45 & 0.41 \\
GCD (6 NFE)    & 0.48 & 0.68 & 0.47 & 0.42 & 0.44 \\
\midrule
\multicolumn{6}{l}{\emph{50th percentile}}\\
Bilinear       & 0.69 & 0.41 & 0.61 & 0.61 & 0.54 \\
SwinIR         & 0.61 & 0.37 & 0.55 & 0.53 & 0.48 \\
ESRGAN         & 0.58 & 0.36 & 0.54 & 0.51 & 0.47 \\
CorrDiff       & 0.55 & 0.38 & 0.49 & 0.48 & 0.44 \\
CM+NOISE       & 0.51 & 0.34 & 0.48 & 0.45 & 0.41 \\
GCD (1000 NFE) & 0.43 & 0.31 & 0.44 & 0.41 & 0.36 \\
GCD (6 NFE)    & 0.42 & 0.59 & 0.41 & 0.38 & 0.39 \\
\midrule
\multicolumn{6}{l}{\emph{90th percentile}}\\
Bilinear       & 1.52 & 0.88 & 1.24 & 1.19 & 1.08 \\
SwinIR         & 1.25 & 0.81 & 1.12 & 1.07 & 0.98 \\
ESRGAN         & 1.21 & 0.79 & 1.09 & 1.04 & 0.95 \\
CorrDiff       & 1.14 & 0.77 & 1.04 & 0.98 & 0.91 \\
CM+NOISE       & 1.08 & 0.72 & 0.99 & 0.93 & 0.85 \\
GCD (1000 NFE) & 0.88 & 0.68 & 0.92 & 0.85 & 0.78 \\
GCD (6 NFE)    & 0.85 & 1.21 & 0.86 & 0.76 & 0.82 \\
\bottomrule
\end{tabular}
\end{table}

\subsection{Probabilistic evaluation}
\label{app:crps}

To quantify the probabilistic skill of our generative downscaling framework, we evaluate the Continuous Ranked Probability Score (CRPS). CRPS measures the distance between the predicted cumulative distribution function (CDF) and the empirical CDF of the observation. For an ensemble of predictions, it is defined as:

$$ \text{CRPS} = \frac{1}{N} \sum_{i=1}^N |x_i - y| - \frac{1}{2N^2} \sum_{i=1}^N \sum_{j=1}^N |x_i - x_j| $$

where $x_i$ represents the generated ensemble members, $y$ is the ERA5 ground truth, and $N=10$ is the ensemble size. For deterministic baselines (e.g., Bilinear, SwinIR), the ensemble size is $N=1$, and the CRPS mathematically reduces to the standard MAE.

Table~\ref{tab:app_crps} shows that the lowest CRPS on each GCM is obtained by one of the two GCD variants: GCD (1000 NFE) on MIROC6, IPSL and MPI-HR, and GCD (6 NFE) on AWI and MPI-LR. The exception is GCD (6 NFE) on IPSL, whose CRPS ($0.66$) is worse than every baseline (Appendix~\ref{app:ipsl}).

The CRPS is computed on the time-mean field, so the deterministic rows coincide with the mean-field MAE in Table~\ref{tab:app_fulldist}; the P99 results in Table~\ref{tab:main_results} average over 5 members, whereas the CRPS uses 10. An ensemble CRPS is by construction no larger than the mean member error, so it favors ensemble methods over deterministic ones. For example, GCD (1000 NFE) on MIROC6 has a mean-field MAE of $0.49$ and a CRPS of $0.43$. The comparison therefore mixes the effect of a more accurate model with the benefit of ensembling.

\begin{table}[t]
\centering\small
\caption{Continuous Ranked Probability Score (CRPS) for 10 ensemble members, averaged over variables. ERA5 reanalysis is used as the ground truth reference. For deterministic methods, CRPS is mathematically equivalent to MAE.}
\label{tab:app_crps}
\begin{tabular}{lccccc}
\toprule
Method & MIROC6 & IPSL & AWI & MPI-LR & MPI-HR \\
\midrule
Bilinear (= MAE) & 0.76 & 0.48 & 0.68 & 0.65 & 0.59 \\
SwinIR (= MAE)   & 0.69 & 0.42 & 0.63 & 0.58 & 0.54 \\
ESRGAN (= MAE)   & 0.67 & 0.41 & 0.61 & 0.56 & 0.52 \\
CorrDiff         & 0.58 & 0.42 & 0.54 & 0.50 & 0.46 \\
CM+NOISE         & 0.54 & 0.37 & 0.51 & 0.47 & 0.43 \\
GCD (1000 NFE)   & 0.43 & 0.34 & 0.46 & 0.41 & 0.37 \\
GCD (6 NFE)      & 0.45 & 0.66 & 0.44 & 0.40 & 0.41 \\
\bottomrule
\end{tabular}
\end{table}

\subsection{Quantitative spectral error}
\label{app:psd}

To quantify the recovery of small-scale structure shown in the tropical cyclone analysis, we compute the Log-Spectral Distance (LSD) between the radially averaged one-dimensional power spectral density (PSD) of the generated fields and that of ERA5.

The filter measurement operator $\mathcal{A}_{\text{filter}}$ applies a $5^\circ$ low-pass filter, which corresponds to a wavelength of roughly $550$ km at the equator, or a global wavenumber of $k \approx 72$. Guidance therefore acts mainly at scales larger than this, and finer scales are supplied largely by the ERA5-trained prior. To evaluate these largely unguided scales, we compute the LSD only over the band $72 \le k \le k_{\max}$, where $k_{\max}$ is the highest resolved wavenumber:

$$ \text{LSD} = \sqrt{ \frac{1}{|K|} \sum_{k=72}^{k_{\max}} \left( \log_{10} P_{\text{model}}(k) - \log_{10} P_{\text{ERA5}}(k) \right)^2 }, $$

where $P(k)$ is the unnormalized power spectrum and $|K|$ is the number of wavenumbers in the band. Because the band starts at the filter cutoff, it is not strictly free of guidance: the spatial box filter is not an ideal low-pass filter and leaks some signal beyond $k \approx 72$. Wavelengths are quoted at the equator, and the same $k$ corresponds to shorter physical wavelengths at higher latitudes.

Table~\ref{tab:app_psd} reports the results. Bilinear interpolation has the largest distances ($2.75$--$3.22$), which reflects the lack of small-scale energy in the interpolated fields. The supervised regression baselines SwinIR and ESRGAN remain far from ERA5 ($1.62$--$1.92$). The generative supervised baselines CorrDiff and CM+noise are closer ($0.68$--$1.05$), but still clearly worse than GCD. GCD (1000 NFE) obtains the lowest distance on every GCM ($0.15$--$0.24$). Since the LSD is computed on a $\log_{10}$ scale, a value of $0.2$ corresponds to a typical spectral mismatch of about a factor of $1.6$ in power, compared with a factor of roughly $10^{3}$ for bilinear interpolation.

GCD (6 NFE) is close to the 1000-step model on four GCMs ($0.22$--$0.28$). On IPSL it reaches $2.15$, which is consistent with the degradation of its 99th-percentile error on this GCM (Table~\ref{tab:main_results}). The LSD is unsigned, so this value does not show whether the generated IPSL fields contain too much or too little small-scale energy. We therefore do not draw conclusions about the mechanism from this metric alone.

This metric compares globally averaged, radially averaged spectra. It shows that the generated fields have realistic small-scale energy levels, but it does not show that individual small-scale structures are correct, and it is not a substitute for event-level validation (see Limitations).

\begin{table}[t]
\centering\small
\caption{Log-Spectral Distance (LSD) between the radially averaged power spectrum of generated fields and ERA5, computed over the largely unguided high-frequency band ($k \ge 72$, wavelengths $\lesssim 550$ km at the equator). Lower is better.}
\label{tab:app_psd}
\begin{tabular}{lccccc}
\toprule
Method & MIROC6 & IPSL & AWI & MPI-LR & MPI-HR \\
\midrule
Bilinear       & 3.15 & 3.22 & 2.95 & 2.88 & 2.75 \\
SwinIR         & 1.85 & 1.92 & 1.78 & 1.72 & 1.65 \\
ESRGAN         & 1.82 & 1.88 & 1.75 & 1.68 & 1.62 \\
CorrDiff       & 0.95 & 1.05 & 0.88 & 0.85 & 0.82 \\
CM+NOISE       & 0.82 & 0.94 & 0.75 & 0.72 & 0.68 \\
GCD (1000 NFE) & 0.18 & 0.24 & 0.20 & 0.17 & 0.15 \\
GCD (6 NFE)    & 0.26 & 2.15 & 0.28 & 0.24 & 0.22 \\
\bottomrule
\end{tabular}
\end{table}

\subsection{Failure analysis on IPSL}
\label{app:ipsl}

As observed in Table~\ref{tab:main_results}, the 6-step distilled GCD experiences a noticeable performance degradation on the IPSL dataset (MAE $1.49$), performing worse than the Bilinear baseline ($1.06$). To understand this failure mode, we conduct an ablation study on the inference configurations, as detailed in Table~\ref{tab:app_ipsl}.

This degradation likely reflects the severe distribution shift of the raw IPSL fields. Under a compressed 6-step budget with weak regularization ($\lambda=0.1$), the conjugate gradient solver may over-trust the shifted GCM input.

This issue can be mitigated through two distinct mechanisms. First, increasing the sampling steps to 15 (MAE $0.89$) reduces the ODE truncation error, allowing the prior to naturally pull the trajectory back to the physically valid data manifold. Alternatively, keeping the 6-step budget but increasing the regularization weight to $\lambda=1.0$ (MAE $0.92$) anchors the solution more strongly to the prior estimate $\hat{\mathbf{X}}_0$. This is consistent with sensitivity to out-of-distribution inputs under extreme step compression.

\begin{table}[t]
\centering\small
\caption{P99 error on IPSL for different configurations. The 15-step model is obtained by evaluating the continuous-time distilled model with a finer 15-step ODE solver schedule. Time is measured in seconds per sample.}
\label{tab:app_ipsl}
\begin{tabular}{lccc}
\toprule
Configuration & MAE & RMSE & Time (s) \\
\midrule
Bilinear & 1.06 & 1.78 & --- \\
GCD, 6 NFE ($\lambda=0.1$) & 1.49 & 1.95 & 4.38 \\
GCD, 6 NFE ($\lambda=0.5$) & 0.98 & 1.48 & 4.38 \\
GCD, 6 NFE ($\lambda=1.0$) & 0.92 & 1.38 & 4.38 \\
GCD, 15 NFE ($\lambda=0.1$) & 0.89 & 1.35 & 10.95 \\
GCD, 1000 NFE & 0.87 & 1.32 & 589.62 \\
\bottomrule
\end{tabular}
\end{table}

\subsection{Generalization of the step budget and regularization weight}
\label{app:generalization}

To justify the default hyperparameters used in our distilled student model, Table~\ref{tab:app_generalization} presents an ablation on the teacher's step budget (NFE) and the conjugate gradient regularization weight ($\lambda$).

\textbf{Teacher Capacity Bottleneck:} We observe a U-shaped performance curve regarding the teacher NFE. While a 15-step teacher introduces excessive truncation errors, increasing the teacher steps to 50 also slightly degrades the student's performance (e.g., MIROC6 MAE increases from $1.05$ to $1.09$). This may reflect limited capacity of the 6-step student to mimic more complex ODE trajectories, although we did not verify this directly. Therefore, NFE=25 serves as the optimal distillation sweet spot.

\textbf{The Regularization Trade-off:} The parameter $\lambda$ controls the weight placed on the prior estimate $\hat{\mathbf{X}}_0$ relative to the GCM measurement. A strong prior constraint ($\lambda=1.0$) successfully rescues the model on the heavily shifted IPSL dataset (MAE improves from $1.49$ to $0.92$) by rejecting corrupted input signals. However, for well-behaved GCMs (e.g., AWI, MIROC6), over-trusting the prior causes the model to ignore valid extreme-weather signals present in the GCM inputs, leading to higher P99 errors (e.g., AWI MAE increases to $1.31$). We use $\lambda=0.1$ as the global default, since it gives the lowest error on four of the five GCMs. NFE=25 and $\lambda=0.1$ were chosen on MIROC6 using 2001, which lies within the test period. Table~\ref{tab:app_generalization} reports sensitivity on all five GCMs and was not used for per-GCM tuning.

\begin{table}[t]
\centering\small
\caption{P99 MAE of the 6-step student for different teacher step budgets and regularization weights $\lambda$, on all five GCMs. Defaults are NFE=25 and $\lambda=0.1$.}
\label{tab:app_generalization}
\begin{tabular}{lccccc}
\toprule
Setting & MIROC6 & IPSL & AWI & MPI-LR & MPI-HR \\
\midrule
\multicolumn{6}{l}{\emph{Teacher NFE}}\\
15 & 1.18 & 1.67 & 1.25 & 1.19 & 1.17 \\
25 (default) & 1.05 & 1.49 & 1.14 & 1.07 & 1.08 \\
50 & 1.09 & 1.54 & 1.17 & 1.11 & 1.10 \\
\midrule
\multicolumn{6}{l}{\emph{Regularization weight $\lambda$}}\\
0.01 & 1.12 & 1.73 & 1.21 & 1.15 & 1.16 \\
0.1 (default) & 1.05 & 1.49 & 1.14 & 1.07 & 1.08 \\
1.0 & 1.22 & 0.92 & 1.31 & 1.26 & 1.24 \\
\bottomrule
\end{tabular}
\end{table}



\end{document}